\RequirePackage{fix-cm}
\documentclass[twocolumn]{svjour3}          
\smartqed  

\def\probP{\mathcal{P}}
\def\FC{\mathcal{F}}
\def\DC{\mathcal{D}}

\def\AC{\mathcal{A}}
\def\CC{\mathcal{C}}
\def\RF{\mathcal{RF}}
\def\AWC{\AC}
\usepackage{graphicx}
\usepackage[inline]{enumitem}
\usepackage{amsmath}
\usepackage{acronym}
\usepackage{multirow}
\usepackage{hhline}
\usepackage{algorithm}
\usepackage{algpseudocode}
\usepackage{algpascal}
\usepackage{makecell}
\alglanguage{pseudocode}
\usepackage{changes}
\usepackage{subfig}
\usepackage{color}
\definechangesauthor[name={fabrizio}, color=red]{fabrizio}
\usepackage{amsmath,amssymb} 

\usepackage{epstopdf,url}
\acrodef{QFU}[QFU]{Querying by Frequent Uncertainty}
\acrodef{HRQ}[HRQ]{Highest Risk Querying}
\acrodef{SSSL}[SSSL]{Stochastic Semi-Supervised Learning}
\acrodef{CSSSL}[CSSSL]{Controlled Stochastic Semi-supervised Learning}
\acrodef{PCA}[PCA]{Principal Component Analysis}
\acrodef{MSE}[MSE]{Mean Squared Error}
\acrodef{JPD}[JPD]{Class Conditional Probability Distribution}
\acrodef{LOF}[LOF]{Local Outlier Factor}
\newcommand{\Rea}{\mathbb R}

\begin{document}

\title{Streaming Active Learning Strategies for Real-Life Credit Card Fraud Detection: Assessment and Visualization
}


\author{Fabrizio Carcillo \and
Yann-A\"el Le Borgne \and
Olivier Caelen \and
Gianluca Bontempi}


\institute{
F. Carcillo, Y. Le Borgne and G. Bontempi \at
Machine Learning Group\\Computer Science Department\\
Faculty of Sciences ULB\\
Universit\'e Libre de Bruxelles\\
Brussels, Belgium\\
\email{{fcarcill, yleborgn and gbonte} @ulb.ac.be} \and
O. Caelen \at
R\&D, Worldline\\
Brussels, Belgium\\
\email{olivier.caelen @worldline.com} }

\date{Received: 13 November 2017 / Accepted: 22 March 2018}

\maketitle

\begin{abstract}
Credit card fraud detection is a very challenging problem because of the specific nature of transaction data and the labeling process.  The transaction data is peculiar because they are obtained in a streaming fashion, they are strongly imbalanced and prone to non-stationarity. The labeling is the outcome of 
an active learning process, as every day human investigators contact only a small number of cardholders (associated to the riskiest transactions)
and obtain the class (fraud or genuine) of the related transactions.
An adequate selection of the set of cardholders is therefore crucial for an efficient fraud detection process.

In this paper, we present a number of active learning strategies and we investigate their fraud detection accuracies. We compare different criteria (supervised, semi-supervised and unsupervised) to query unlabeled transactions.
Finally, we highlight the existence of an exploitation/exploration trade-off for active learning in the context of fraud detection, which has so far been overlooked in the literature. 
\footnote{This paper is an extension version of the DSAA'2017 Application Track paper titled:
``An Assessment of Streaming Active Learning Strategies for Real-Life Credit Card Fraud Detection'' \cite{carcillo2017active}}
\keywords{active learning \and fraud detection \and selection bias \and semi-supervised learning}
\end{abstract}
\begin{sloppypar}
\section{Introduction}

\label{section:intro}

The use of machine learning for credit card fraud detection requires to address a number of challenges. Some of them are related 
to the data distribution, notably the class imbalance of the training set
(many more genuine transactions than fraudulent ones), the non-stationarity of the phenomenon (due to changes in the behavior of customers as well as in fraudsters),
the large dimensionality and the overlapping classes (while fraudsters try to emulate cardholders behavior, genuine behaviors of cardholders might look strange or anomalous). 

The labeling process is constrained, as every day human investigators may contact only a small number of cardholders associated with the riskiest transactions and obtain the class (fraud or genuine) of the related transactions.
The high cost of human labour, for assessing the transaction labels, leads to the labeling bottleneck~\cite{bhattacharyya2011data}.

In this context, an automatic Fraud Detection System (FDS) should support the activity of the investigators by letting them focus on the transactions with the highest fraud probability.
From the perspective of the transactional service company, this is crucial  in order to reduce the costs of the investigation activity and to retain the customer confidence.
From a machine learning perspective it is important to keep an adequate balance between \emph{exploitation} and \emph{exploration}, i.e. between the short-term needs 
of providing  good alerts to investigators, and the long-term goal of maintaining a high accuracy of the system (e.g. in the presence of concept drift).

The issue of  labeling the most informative data by minimizing the cost has been extensively addressed by \emph{active learning} which can be considered
as a specific instance of \emph{semi-supervised learning}~\cite{chapelle2009semi,settles2010active}, the domain studying how unlabeled and labeled data can both contribute to a better learning process.
The Active Learning (AL) literature have proposed a number of techniques to select, within a large set of unlabeled instances, the most informative to label. 
AL can be typically described by the iteration of three steps~\cite{settles2010active}:
\begin{enumerate}
  \item \textit{Selection}: given a query budget of size $k$, a  model is used to choose which  $k$ data points, once labeled, could better describe the data generating process;
  \item \textit{Querying}: selected  points are submitted to an \textit{oracle} (in our case an investigator) for labeling;
  \item \textit{Training}: the labeled data points are used to train and update the model;
\end{enumerate}
The \textit{Selection} and \textit{Querying} steps differentiate active learning from conventional (passive) learning, which is limited to the \textit{Training} step.
Note also that, in order to bootstrap the procedure, a random initialization or an unsupervised model is used \cite{settles2010active}. 

A first classification of AL methods can be done according to the model used for the  \textit{Selection} step.
In this paper, we will distinguish between methods using only supervised information and methods integrating other sources of information, notably unsupervised or semi-supervised.
Additionally, according to the nature of the dataset, we can distinguish between two main AL approaches: \emph{pool-based} and \emph{streaming} AL.
In pool-based AL, the algorithm performs queries in the same set of unlabeled points, while in stream-based AL, the set of unlabeled data points is periodically updated.
The accuracy of a pool-based AL classifier is expected to grow in time, since more and more labeled data points, from the original dataset, are used for the \textit{Training}.
This is not always true in the case of the streaming approach, since data received in different periods may differ significantly (e.g. concept drift).

Though AL is widely addressed in the literature~\cite{cohn1994improving,lewis1994sequential}, few articles mention the aspects of credit card fraud detection (Section~\ref{subsection:actlea}), notably the class imbalance \cite{ertekin2007learning,zhu2007active} (in our case study approximately only 0.2\% of  transactions\footnote{
Though some papers on fraud detection present datasets with still lower rates (0.01\% in \cite{dorronsoro1997neural}, 0.005\% in \cite{bhattacharyya2011data}, 0.02\% in \cite{wei2013effective} and 0.004\% in   \cite{sahin2013cost}) our dataset is inline with other recent works on fraud detection (\cite{van2015apate},\cite{jurgovsky2018sequence} and~\cite{seeja2014fraudminer}  have a class imbalance rate of 0.8\%, 0.5\% and 0.4\% respectively).}
 are fraudulent), the restricted labeling budget \cite{schohn2000less,vijayanarasimhan2010far} and the specific nature of the assessment \cite{settles2008multiple}.
It is worth to remark that in credit card fraud detection, though the observations are at the level of transactions, the ultimate goal is to detect fraudulent cards.
There exists a one-to-many relation between cards and transactions.

A peculiarity of fraud detection is that the labeling and the assessment phase are coincident and consequently strongly dependent. 
For the transactional service provider it is important that the investigations are as successful as possible (i.e. low false positive rate).
This means that the accuracy of a FDS is measured in terms of precision over the top $k$ alerted credit cards~\cite{carcillo2017InfFus,dal2017TNNLS}. 
This is not always the case in other AL tasks where the labels of the $k$ queried points are not directly related to the accuracy of the training process. 
In a FDS, it is not only important to minimize the labeling cost  but also that this labeling allows to discover as many frauds  as possible.
The nature and the intensity of the exploration step has a strong impact on the final accuracy of detection and, accordingly, the set of state-of-the-art AL strategies which are effective in practice is much more limited than expected. In other terms no real-life FDS can afford a totally random labeling process since this would necessarily imply an unacceptable short-term random performance. This exploitation/exploration trade-off inherent to fraud detection, to the best of our knowledge, has not been addressed in the research literature.

The contributions of this article are: i) a taxonomy of streaming AL strategies (and a number of their variants) for credit card fraud detection, ii) an extensive comparison of these techniques for the detection of both fraudulent transactions and cards, iii) an  experimental assessment on the basis of a massive set of 12 million transactions  in terms of real-life criteria (defined by our industrial partner, Worldline, a leader company in transactional services) and iv) a two-dimensional visualisation of the effects of  active learning on the distribution of the training set.
The outcome is an original analysis of the exploitation/exploration trade-off in the context of a real-world FDS.
In particular, we expect that visualization represents a valuable insight on the evolution of the training set during active learning.
In fact, though the rationale of active learning is explicit, it is not always evident to understand how active learning modifies the distribution of the training set, notably in a task, like credit card fraud detection, characterized by large noise, nonlinearity and non separability.
Also, we expect that potential bias of AL sampling strategies are easier to assess in a visual setting.

The work is organized as follows.
Section \ref{section:rw} presents the related state-of-the-art.
Section \ref{section:ps} provides the general outline of our Fraud Detection System.
Section \ref{section:dv} introduces the visualization technique used throughout the manuscript.
Section \ref{section:str} discusses a number of AL strategies (as well as possible variants) for dealing with streaming credit card transactions.
Finally, Section \ref{section:exp} presents an extensive experimental session based on a real stream of transactions.

\section{Related work}
\label{section:rw}

In this section, we will present some important state-of-the-art works in fraud-detection. 
The review of conventional passive learning for fraud detection is presented in the Section \ref{subsection:outdet}, while active learning techniques and their application to fraud detection are introduced in Section \ref{subsection:actlea}.
\subsection{Passive learning for fraud detection}
\label{subsection:outdet}
Credit card fraud detection belongs to the largest domain of outlier detection~\cite{aggarwal2015outlier,chandola2009anomaly,pimentel2014review} also called anomaly or novelty detection. The main approaches to anomaly detection are:
\begin{enumerate}
  \item Classification based: these supervised techniques make the assumption that a classifier of anomalies can be learnt from a training set and that the test set distribution is not significantly different from the training one.  
Several supervised machine learning algorithms have been discussed in  literature~\cite{sethi2014revived,shimpi2015survey,zareapoor2015application}. 
In most cases an optimal detection can be obtained by combining multiple supervised machine learning techniques. For instance, Wei et al. \cite{wei2013effective} introduce ContrastMiner, a fraud detection framework which combine the use of contrast pattern mining, neural network and decision forest to have a high precision in the detection.
In our previous research we have used and assessed several binary classifiers for fraud detection~\cite{dal2017TNNLS,dal2014learned}. 
Also one-class classifiers, like one-class SVM and Isolation Forest, belong to this category.
One-class SVM~\cite{scholkopf2000support} fits a boundary around the known set of normal points, and then classifies as outlier everything which stands over the boundary. The Isolation Forest~\cite{liu2008isolation} uses the length of the path between the root and the leaves in the trees of a Random forest as outlier score.

  \item Nearest neighbor based: the rationale is that the distribution of the neighbors (or local density) of a point characterizes the genuine or anormal nature of a point. In particular, normal instances stand in dense areas while outliers are located far from dense areas. \ac{LOF} is a well-known density/neighbour based technique proposed by Breunig et al. in \cite{breunig2000lof}. Variants of LOF have been proposed in~\cite{kriegel2009loop,ren2004rdf,tang2002enhancing,zhang2009new}.
  \item Clustering based: once the training set is properly clustered, outliers are expected to be located much farther away from the clusters center than normal ones.
  \item Statistically based: they require the estimation of the multivariate distribution of data and return a score of outlierness which is inversely proportional
  to the density of a point. Gaussian Mixture Models (GMM)~\cite{drews2013novelty,ilonen2006gaussian,li2016anomaly,zhang2017detection}  are commonly used in this context because of their flexible semi-parametric nature. Hidden Markov Models have been used in~\cite{srivastava2008credit} for credit card outlier detection.  Bolton et al. \cite{bolton2001unsupervised} presented two 
  statistical approaches which rely on monitoring the anomalous behavior of a cardholder compared to a group of peers (peer group analysis) or to cardholder behavior (break point analysis).

  \item Information theoretic: they rely on the fact that anomalies create irregularities in the dataset and can be detected with information theoretic measures.
  \item Spectral anomaly detection: these techniques transform the original dataset in a way that facilitate the separation of normal instances and outliers.
A commonly used transformation technique is the \ac{PCA}~\cite{pinto2011weighted,palau2011burst,wang2004novel}.
Shuy et al. \cite{shyu2003novel} proposed to combine the effect of the major component and the minor component.
While the major component is used to find extreme values, the minor component highlights those observations which do not conform to the normal correlation structure. 
\item Outlier detection combined with feature selection: unsupervised outlier detection has been recently combined with feature selection.
  Pang et al. proposed a filter-based feature selection framework for unsupervised outlier selection \cite{pang2016unsupervised}.
  This approach improved the outlier detection rates while substantially reducing the number of features by 46\% (on average).
  The same authors proposed a novel wrapper-based outlier detection framework~\cite{pang2017learning} to iteratively optimize the feature subset selection and the outlier score. The results showed that the framework improved not only the AUC-ROC, but also the precision over the top-n ranked instances.
\end{enumerate}




\subsection{Active learning}
\label{subsection:actlea}

Unlike passive learning, active learning modifies the size and nature of the training set
by choosing from a unlabeled set a subset of points whose label is expected to improve the classifier. The AL setting is particularly
promising in fraud detection because of the cost and the delay related to the labeling of instances.
However, the adaptation of active learning to the specific characteristics of fraud detection data has only been partially addressed in the research literature.
Fan et al.~\cite{fan2004active} carried out an empirical analysis on a fraud detection dataset to assess AL approach in presence of concept drift.
They have focused on the adaptation ability of the AL strategy, but did not address the detection accuracy.
Pichara et al. \cite{pichara2008detection} tested a large scale anomaly detection approach in a synthetic dataset emulating the fraud process.
Their AL schema has been able to detect the whole subset of frauds using a number of queries smaller than a Bayesian Network detection approach.
Multiple tests were repeated using different data-noise levels, and their AL approach consistently outperformed the other techniques.

However, the use a synthetic dataset reduces the impact of these results. It is very difficult to create a reliable and synthetic credit card dataset, since transactions (frauds and genuine) are very diverse and evolve in an unpredictable way.
Van Vlasselaer et al. \cite{van2015afraid} applied active inference, a network-based algorithm, to fraud discovery in social security real data.
They found that committee-based strategies, based on uncertainty, result in a slightly better classification performance than expert-based strategies.
Nevertheless, expert-based strategies are often preferred in order to obtain unbiased training sets from queries.

The relationship between active learning and streaming data, notably the sampling bias issue, is discussed in~\cite{dasgupta2011two}.
The authors showed that in stream-based active learning, the estimated input-output dependency changes over time and depends on the instances previously queried.
Since those instances are typically selected next to the class decision boundary of the classifier, this may lead to a biased representation of the underlying data distribution.
AL and concept drift is also addressed in~\cite{vzliobaite2011active}. Here, the authors stressed how concept drift may be missed in regions far from where AL queries normally take place (e.g. boundary regions between classes).
The authors showed that techniques based on classical uncertainty sampling favor close concept drift adaptation while techniques based on random sampling 
are more effective in dealing with remote concept drift. The best performing techniques can strongly depend on the characteristics of the data and the size of the query budget.

The issue of sampling bias is specifically discussed in \cite{jacobusse2016selection}. Jacobusse and Veenman present multiple solutions to tackle the sampling bias on highly imbalanced datasets and screening applications.
In such conditions, a small group of targets need to be detected among the non-targets vast majority.
A random selection would lead to a very poor detection.
So, usually an expert knowledge driven selection is preferred (medical screening application, law enforcement, screening of job applicants, ...).
The authors emphasize that this selection is prone to suffer of a strong bias towards the previous knowledge of the expert, causing the classifier to be trained on a non representative dataset. A similar problem is faced in fraud detection, since only a small subset of transactions can be labeled in the short term and the selection of this subset corresponds to the riskiest credit card transactions.

The integration of AL and semi-supervised learning is discussed in Xie and Xiong \cite{xie2011stochastic}.
They introduced a \ac{SSSL} process to infer labels in case of large imbalanced datasets with small proportion of labeled points.
The approach relies on the consideration that since the number of unlabeled points is huge and the minority class is rare, the probability of making a wrong majority assignment is very low.
Consequently, they proposed the assignment of the majority class to random selection of points and adopted it with success
in the context of a data competition.

Finally, an original approach that may be used to deal with the one-to-many relationships between cards and transactions is discussed in~\cite{settles2008multiple}.
They present an AL approach for multiple-instance problems where instances are organized into \emph{bags}.
Typical examples of multiple-instance problems are found in text classification and content-based image retrieval.
In these type of problems a bag is said to be positive if it includes at least one instance which is positive, while the  bag is negative if no positive instances are observed in it.

\section{The fraud detection system classifier}
\label{section:ps}
Let us consider a Fraud Detection System (FDS) whose goal is to detect automatically frauds in a stream of transactions.
Let $x \in \Rea^n$ be the vector coding the transaction (e.g. including features like the transaction amount, the terminal) and $y \in \{+,-\}$ the associated label, where $+$ denotes a fraud and $-$ a genuine transaction.  
A detection strategy needs a measure of risk (score) associated to any transaction.
In a machine learning approach this score is typically provided by the estimation of the a posteriori probability $\probP_{\CC}(+|x)$  returned by a classifier $\CC$.
We consider a streaming setting where unlabeled transactions arrive one at a time or in small batches.

The FDS goal is to raise every day a fixed and small number of $k$ alerts. In our industrial case study, $n=32$ and  $k$ is set to $100$ on the basis of cost and work organization considerations.

The issuing of those alerts has two consequences: the trigger of an investigation and the consequent labeling of the associated transactions.
The outcome of the investigation determines both the success rate of the FDS and the new set of labeled transactions.

We will present in section~\ref{section:exp} two levels of experimental validations: the first concerns the detection of fraudulent transactions, while the second focuses on fraudulent cards. In the first experiment, the classifier $\CC$ is implemented by a conventional Random Forest, while in the second, we use a more complex approach (ensemble of classifiers) dictated by the more challenging nature of the detection tasks.
This approach has been presented in~\cite{carcillo2017InfFus,dal2017TNNLS} and consists of the weighted average of two classifiers
\begin{equation}
\label{eq:pat}
\probP_{\CC}(+|x) = w^A\probP_{\DC_{t}}(+|x) + (1-w^A)\probP_{\FC_{t}}(+|x)
\end{equation}

where $\DC_{t}$ and $\FC_{t}$ stand for \textit{Delayed classifier} and \textit{Feedback classifier} respectively and $w^A \in [0,1]$ is the weight controlling the contribution of the two classifiers.
$\DC_{t}$ is implemented as an ensemble of Balanced Random Trees \cite{chen2004using,Rokach2016111} trained on old transactions for which we can reasonably consider the class as known.
$\FC_{t}$ is trained on recently alerted transactions, for which a \textit{Feedback} was returned by investigators. It is therefore alimented by the active learning component of the fraud detection system.
This \textit{Feedback} component is very important to address concept drift.

This architecture is the result of an extensive model selection and assessment procedure which has been discussed in our previous work \cite{carcillo2017InfFus,dal2017TNNLS}.
The aim of this paper is to discuss the impact of different AL strategies, so we will not take into consideration alternative classifier architectures.

\section{Dataset visualization}
\label{section:dv}
\begin{figure}[!t]
\centering
\includegraphics[width=3.2in, height=3.2in]{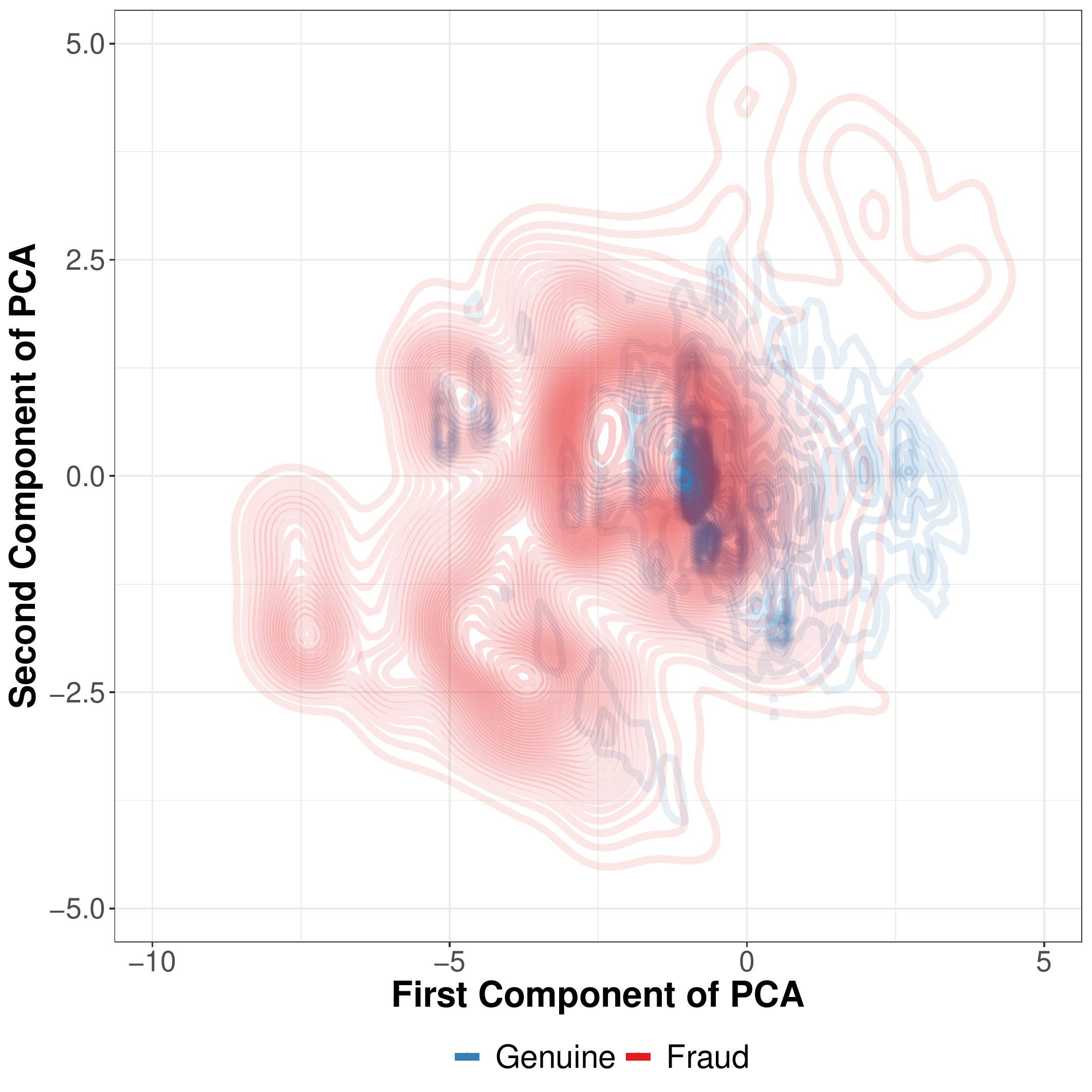}
\hfil
\caption{Class conditional distributions (level curves of Frauds in red and of Genuine in blue) of  transactions in the PC1/PC2 space. The dataset of transactions is collected on 15 consecutive days.
\label{fig:jpdf1}}
\end{figure}

Credit Card Fraud Detection deals with high-volume (millions of transactions) and large dimensionality ($n=32$ in our example) datasets. So, it is difficult to have a visual insight of the data distribution and the nature (e.g. non-separability) of the classification task. As a consequence most papers rely only on predictive accuracy (e.g. ROC curves) to assess the difficulty of the task. Given the dynamic and sampling nature of AL , it is however relevant to visualize the location of the selected query points with respect to the original data distribution in order to better illustrate the differences between alternative strategies. In the following section we will complement the presentation of the different AL strategies with a visual representation of our experimental dataset (details are in Section~\ref{section:exp})  in the space of the two  first Principal Components (denoted by PC1 and PC2). Fig.~\ref{fig:jpdf1} shows the Class Conditional distributions of the two classes (Fraud in red and Genuine in blue) in the PC1/PC2 space.

This visualization provides interesting insights about the distribution of Frauds and Genuine transactions: i)  the two classes appear to be only partially overlapping in the space, ii) the density of the fraudulent set appears to have a higher variance than the genuine set and iii) as far as the first principal component is concerned, frauds have a distribution skewed to the left.

However, the class-conditional nature plot should not mislead us to the conclusion that the problem is easy to solve.
In fact, given the high imbalance of the classes, a large number of genuine transactions still occur in the left part of the plot (mostly red).

In Section~\ref{section:str} we will use Fig.~\ref{fig:jpdf1} as a template for illustrating the distribution of the queries issued by the different AL strategies taken into consideration.


\begin{table*}[!t]
\fontsize{9}{9}\selectfont
\setlength{\tabcolsep}{1em}
\renewcommand{\arraystretch}{1.5}
\caption{Summary of active learning and semi-supervised strategies described in the paper}
\label{syntstr}
\centering
\begin{tabular}{|c||c|c|}
\hline
\textbf{Id} & \textbf{Strategy} & \textbf{Type}\\
\hline
HRQ & Highest Risk Querying &  Baseline / BL (Section \ref{subsection:strHRQ})\\
\hline
R & Random Querying & \\  
P & Unsupervised (PCA) Querying & \\
U & Uncertainty Querying & \\

M & Mix of Random and Uncertainty Querying &\multirow{-3}{*}{\makecell{Exploratory Active Learning / \\ EAL (Section \ref{subsection:sals})}  }  \\
\hline
SR & SSSL on Random points & \\
SU & SSSL on Uncertain points &\\
SM &  SSSL on Random/Uncertain points &\\
SE & SSSL on points most likely to be genuine & \multirow{-4}{*}{\makecell{Stochastic Semi-Supervised Learning / \\ SSSL (Section \ref{subsection:sal})}}\\
\hline
SR-U & SSSL on Uncertain points + Random Sampling & \\
SR-R & SSSL on Random points + Random Sampling &\\
SR-M & SSSL on Random/Uncertain points + Random Sampling & \multirow{-3}{*}{SSSL + EAL (Section \ref{subsection:sal})}\\
\hline
SRN[$p$] & SR with reduced x\% of Negative Feedback & Modified SSSL (Sections \ref{subsection:sal})\\
\hline
ROS & Random Oversample &\\
SMOTE & SMOTE & \multirow{-2}{*}{Oversample (Section \ref{subsection:os})}\\
\hline
QFU & Querying by Frequent Uncertainty &\\
MF-... & Max combining function &\\
SM-... & Softmax combining function &\\
LF-... & Logarithmic combining function & \multirow{-4}{*}{Multiple Instance Learning (Sections \ref{subsection:mil})}\\
\hline

\end{tabular}
\end{table*}

\begin{algorithm}
\caption{Active Learning process}
\label{alg:process}
\begin{algorithmic}[1] 

\Require $k$ \Comment{total number of alerts}
\Require $q$ \Comment{exploration budget}
\Require $m$ \Comment{SSSL budget}
\Require $D$ \Comment{ initial training set}

\For{any new day}
\State $\CC \gets$ learning$(D)$
\State $inTrx\gets$ unlabeled set
\State $scores\gets$ $\{ \probP_\CC(x), x \in inTrx \}$
\State $sel\gets$ selection of the $k-q$ points with highest risk $scores$ \Comment{ HRQ}
\If {($q>0$)} \Comment{ EAL}
\State $Esel\gets$ $q$ explorative points 
\State $sel\gets \{sel, Esel\}$ 
\EndIf 
\State $queries\gets$ investigator labeling of $sel$

\If {($m>0$) }  \Comment{SSSL}
\State $SSSLset\gets$ $m$ points based on a SSSL criterion
\State $SSSLset\gets$ set label $y(SSSLset)=0$
\State $queries\gets \{queries, SSSLset \}$

\EndIf 

\State $D \gets \{D, queries \}$
\label{row:end}
\EndFor

\end{algorithmic}
\end{algorithm}

\section{Active learning strategies}
\label{section:str}

The rationale  of AL is to select (on the basis of current information) unlabeled training samples which, once labeled, can improve the accuracy.
However, there are two main unknowns concerning the effectiveness of AL in credit card fraud detection.
The first concerns the strong imbalance of the  class distribution: as the selection of adequate queries is the most important step of an AL procedure.
This step should take into account that in such a large imbalanced problem, selecting majority class points will inevitably have a negligible impact on accuracy.
The second concerns the definition of accuracy: measures of detection accuracy are strictly related to the capacity of discovering frauds, i.e. querying minority class samples. This means that an AL strategy for fraud detection requires some specific tuning for being successful.

To illustrate the impact of AL on FDS, we will start by considering a baseline strategy which simply queries the highest risk transactions on the basis of the current classification model. This strategy will be denoted as the Highest Risk Querying (HRQ). 
Thereafter, we will introduce and assess a number of modifications of HRQ according to several principles.
In order to make the comparison easier we will define each AL strategy as an instance of a generic AL strategy detailed in Algorithm~\ref{alg:process}.
The Algorithm requires the specification of three parameters: the budget $k$ of queries (i.e. maximum number of transactions that can be investigated per day), the number of $q$ queries defined for exploration purposes and the number $m$ of unlabeled transactions that can be set as genuine without investigation (see \ref{subsection:sal}).
When $q>0$, the choice of the queries demands a selection criterion which plays a major role in the final accuracy of AL. 
Note that the criteria used by the methods discussed in the following sections can be regrouped in three classes: supervised (i.e. relying on labeled data), unsupervised and semi-supervised.  
The entire list of discussed AL strategies is presented in Table~\ref{syntstr}.

\subsection{Highest risk querying}
\label{subsection:strHRQ}

The idea of \ac{HRQ} is simple: given a classifier $\CC$ and a budget of queries, HRQ returns the unlabeled transactions
with the highest estimated \textit{a posteriori} probability $\probP_{\CC}(+|x_i)$.
HRQ is the most intuitive active learning strategy for our problem if we consider that the final FDS accuracy depends on the amount of minority class querying.
Note that in terms of the pseudocode in Algorithm~\ref{alg:process}, HRQ is obtained by setting $q=0$ and $m=0$.

HRQ is expected to have a positive impact on accuracy by discovering new instances from the minority class and
improving consequently the balance of the training set. 
It has also some drawbacks: since its querying strategy relies on the classifier accuracy, this selection step could be inaccurate especially at the very beginning.

\begin{figure}[!ht]
\centering
\includegraphics[width=3.2in, height=3.2in]{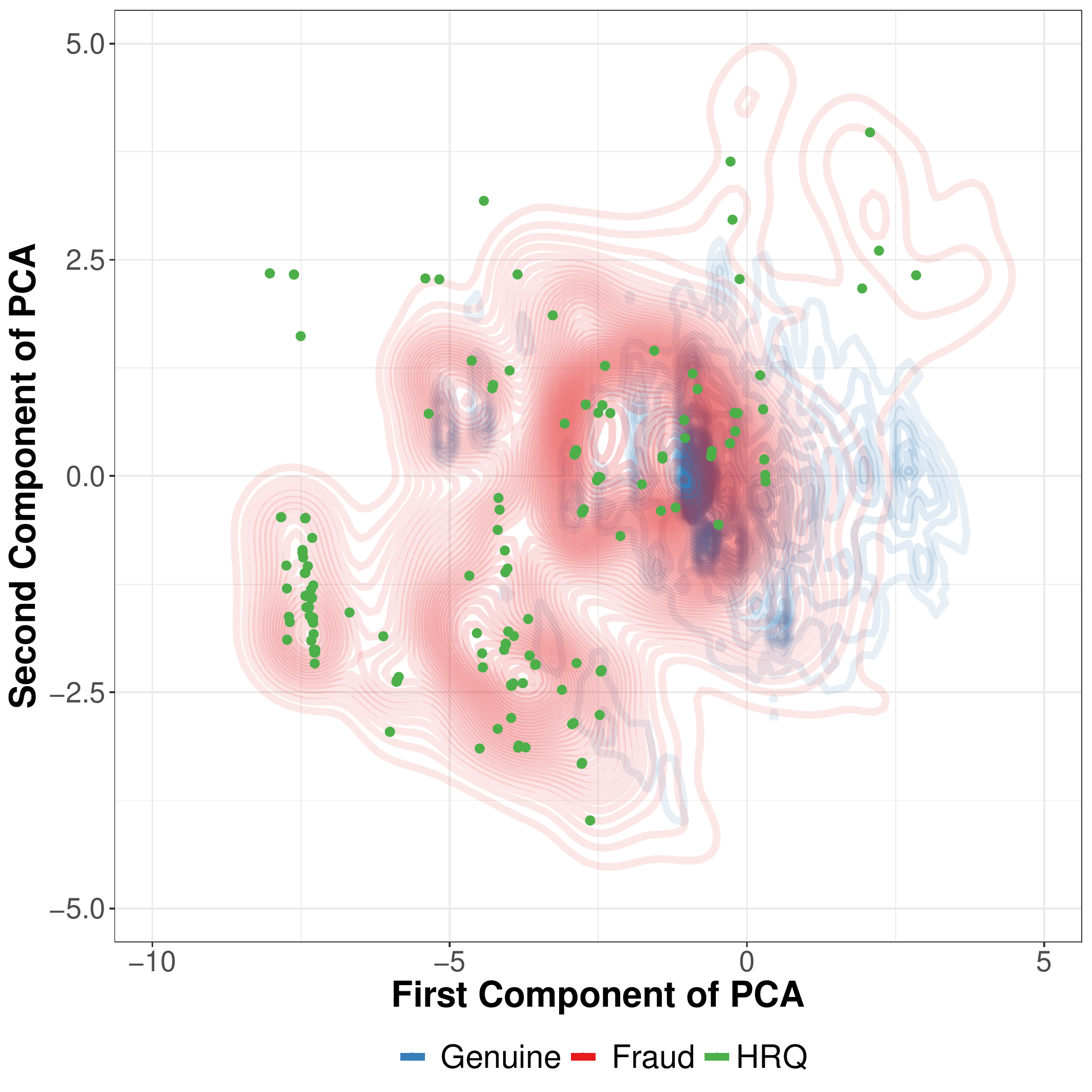}
\hfil
\caption{ Class conditional distributions in the PC1/PC2 space and the set of 200 transactions queried by the HRQ approach.
\label{fig:jpdf2}}
\end{figure}

Fig.~\ref{fig:jpdf2} represents a set of 200 query points returned by \ac{HRQ}. 
As expected, the classifier $\CC$ selects points located in areas where the class conditional density of frauds is high (red areas).
The training of a new classifier using exclusively this subset of points may suffer of selection bias, since the sample is not representative of the distribution of the genuine class.

\subsection{Exploratory active learning}
\label{subsection:sals}
Exploratory Active Learning (EAL) strategies modify \ac{HRQ} by trading exploitation for exploration. The idea is to convert
a subset of the labeling budget in explorative queries. 
The size of the exploration budget is represented by $0<q\leq k$ in Algorithm~\ref{alg:process}.

We may consider a number of exploration techniques for selecting the $q$ exploratory transactions.
The simplest one is random querying (denoted by EAL-R) which consists in choosing randomly the $q$ query points.
This selection is unsupervised and sub-optimal since it may query points for which the classifier is already highly confident about the class.
A less naive unsupervised strategy (denoted by EAL-P) consists in using an unsupervised algorithm (e.g. PCA) to select the $q$ queries.

An  alternative with a supervised selection criterion is represented by uncertainty querying (EAL-U) which returns unlabeled data points for which the current classifier has low confidence~\cite{Lewis94heterogeneousuncertainty}. The selection criterion is therefore supervised: given a binary classifier $\CC$, the uncertainty querying strategy gives priority to  the transactions $x_i$ for which $\probP_{\CC}(+|x_i) \approx 0.5$. 
This value may be affected by class imbalance and an higher threshold may be chosen in the fraud detection setting.
Nevertheless, it is a good practice to balance the dataset before training the classifier.
In the experimental section, we will implicitly refer to a classifier $\CC$ which is trained on a balanced dataset.

\begin{figure}[!ht]
\centering
\includegraphics[width=3.2in, height=3.2in]{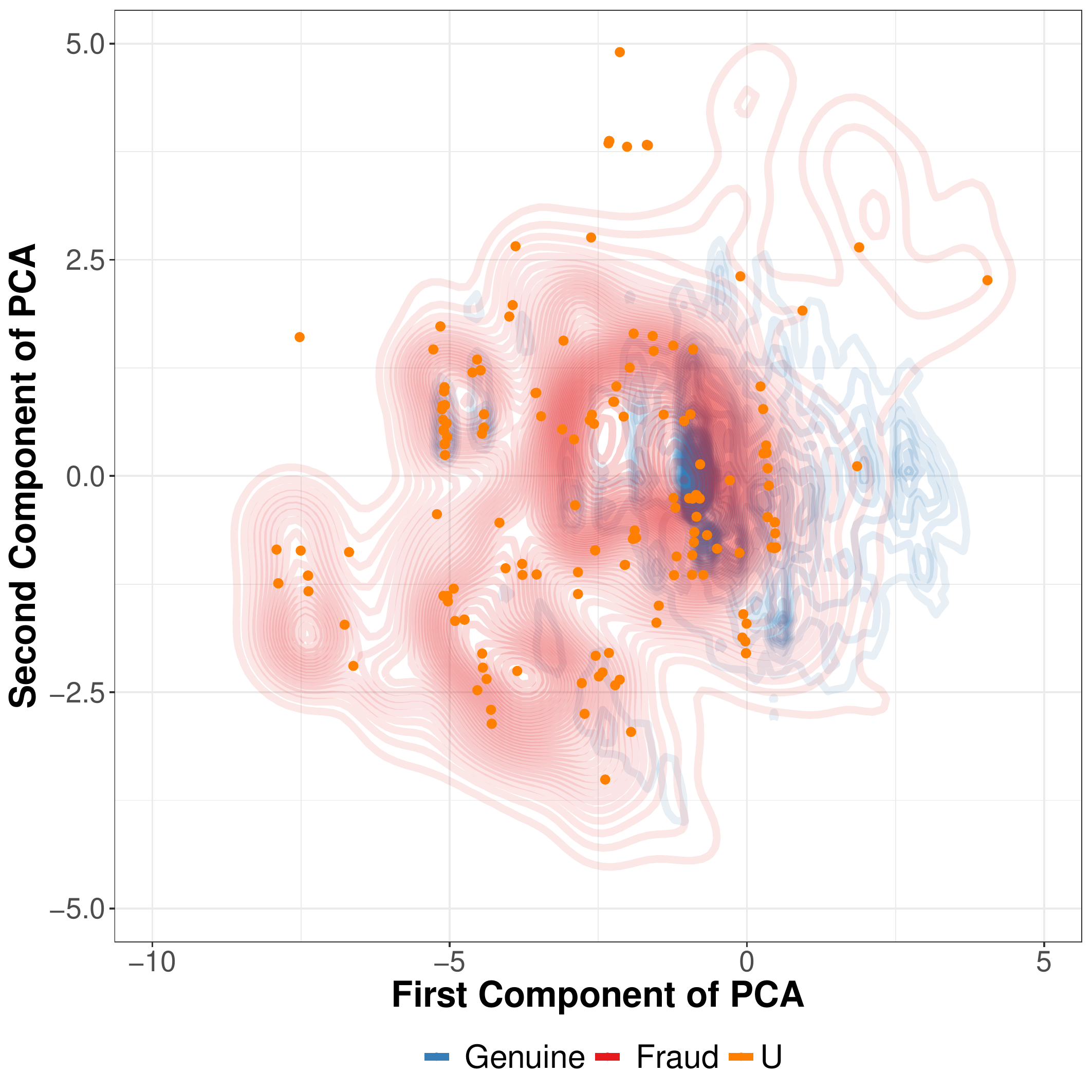}
\hfil
\caption{Class conditional distributions in the PC1/PC2 space and the set of 200 transactions queried by the Uncertainty Sampling approach.
\label{fig:jpdf3}}
\end{figure}

Fig. \ref{fig:jpdf3} presents the visualization of a set of 200 transactions selected by the Uncertainty Sampling approach. 
The transactions are expected to be selected in areas where the classifier $\CC$ is the most uncertain.
As a consequence, selected transactions will lie in areas which present a high density for both classes.

\v{Z}liobait{\.e} et al. \cite{vzliobaite2011active} proposed the mix of the two techniques, uncertainty querying and randomization, to tackle remote concept drift (Section~\ref{section:rw}).
The technique (denoted by EAL-M) consists in querying by uncertainty  most of the points and in querying random points from time to time.

\subsection{Stochastic semi-supervised learning}
\label{subsection:sal}
The \acf{SSSL} strategy has been introduced by Xie and Xiong \cite{xie2011stochastic} to infer labels in case of highly imbalanced datasets with a large number of unlabeled points.
The strategy relies on a simple consideration: since the ratio between the number of frauds and the total number of transactions is very small, the probability of randomly selecting a fraud is very low. 

The resulting AL learning schema is made of four steps:
\begin{enumerate}
  \item \textit{Selection}: the current model is used to annotate all unlabeled transactions with an estimated risk;
  \item \textit{Querying}: the highest risk transactions are labeled by the investigators;
  \item \textit{Majority assumption}: a number of transactions are labeled as genuine by majority assumption; in this paper we explore a number of criteria to attribute the majority class: pure random attribution (SR), uncertainty (SU), mix of randomness and uncertainty (SM) and low predicted risk (SE).
  \item \textit{Training}: the labeled data points, obtained by the previous steps, are used to train/update a supervised model.
\end{enumerate}
Though the selection strategy is supervised as in EAL, SSSL differs in terms of the usage of the current model $\CC$: the predicted risk is not only used to alert
and trigger the investigation but also to label (without investigation) a number of low risk transactions.

\begin{figure}[!t]
\centering
\begin{minipage}{1\linewidth}
\subfloat[]{\includegraphics[width=1.75in, height=2in]{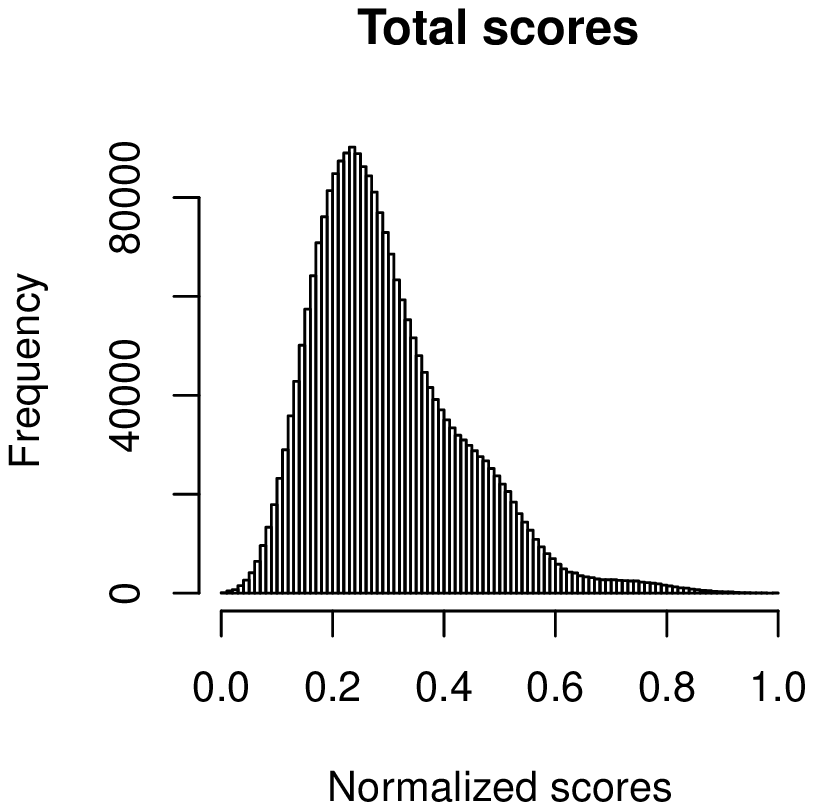}
\label{fig:sc1}}
\subfloat[]{\includegraphics[width=1.75in, height=2in]{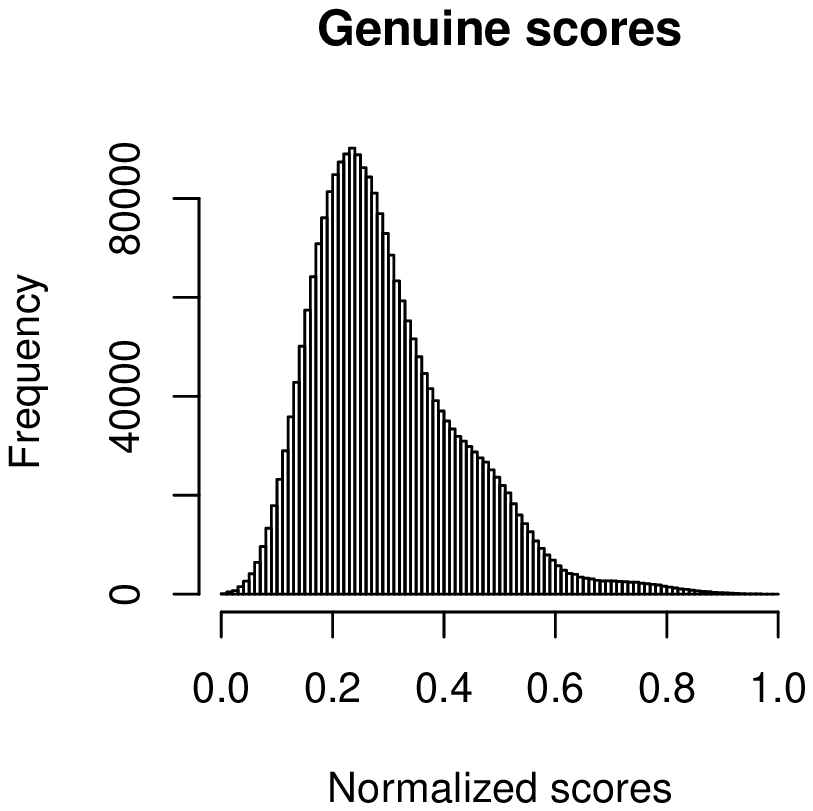}
\label{fig:sc2}}
\end{minipage}\par\medskip
\begin{minipage}{1\linewidth}
\subfloat[]{\includegraphics[width=1.75in, height=2in]{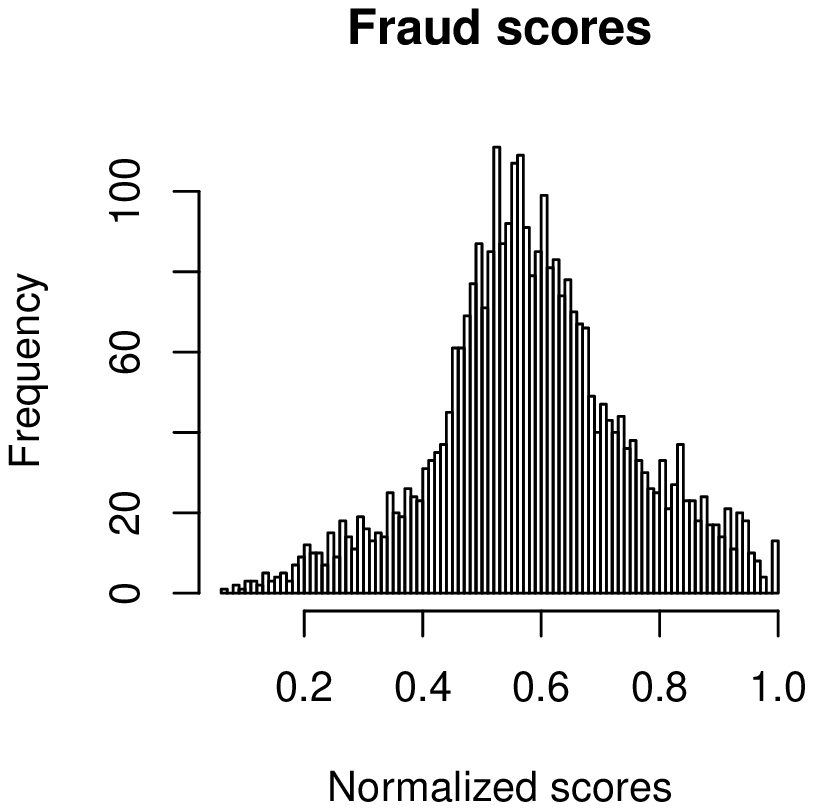}
\label{fig:sc3}}
\subfloat[]{\includegraphics[width=1.75in, height=2in]{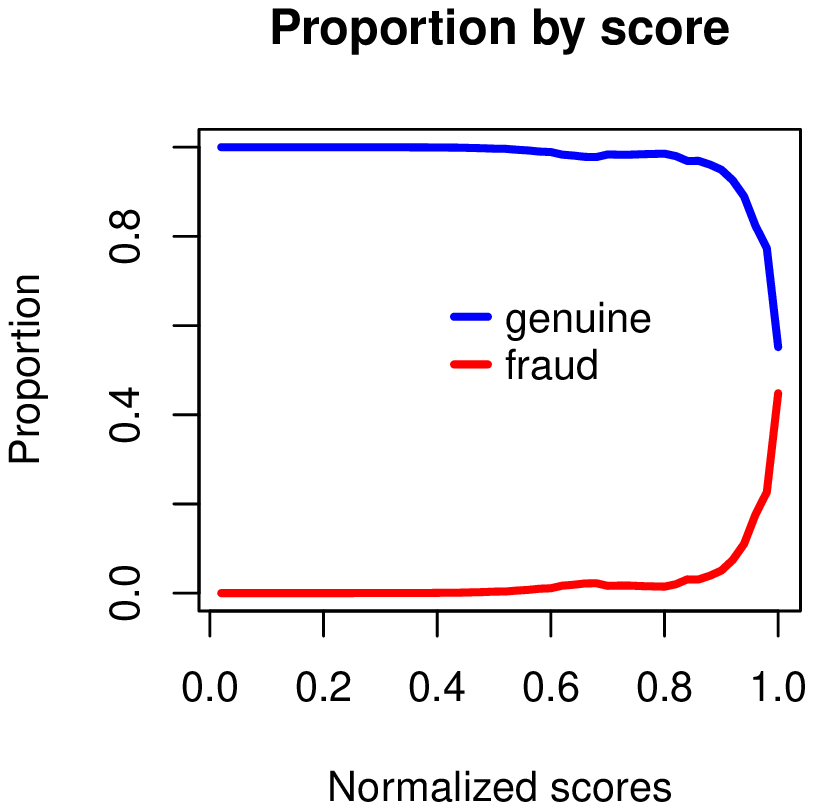}
\label{fig:sc4}}
\end{minipage}
\hfil
\caption{Distribution of the scores obtained by $\probP_{\AWC_{t}}(+|x_i)$ in a range of 15 days involving 2.4 millions of cards for: all the transactions (a), only genuine transactions (b) and only fraudulent transactions (c). In (d), the proportion of genuine and fraudulent transactions is plotted while changing the score obtained by $\probP_{\AWC_{t}}(+|x_i)$.\label{fig:ns}}

\end{figure}

In order to illustrate the reliability of the majority assumption, we report in Fig.~\ref{fig:ns}  the distribution of the scores $\probP_{\CC}(+|x_i)$ over 15 days.
In particular  the histograms (a), (b) and (c)  refer to the score distribution for all transactions, genuine and fraudulent transactions, respectively. The plot (d) represents the proportion of fraudulent and genuine transactions for a given score in the range $[0,1]$. Note that, though the a priori proportion of fraudulent cards in the dataset is 0.13\%,  it becomes 23.33\% for scores higher than 0.95 and 61.90\% for scores beyond 0.99.
Note also that, in the area of maximal uncertainty for $\CC$ (e.g between  0.49 and 0.51), we find only 0.35\% of frauds.

\begin{figure}[!ht]
\centering
\includegraphics[width=3.2in, height=3.2in]{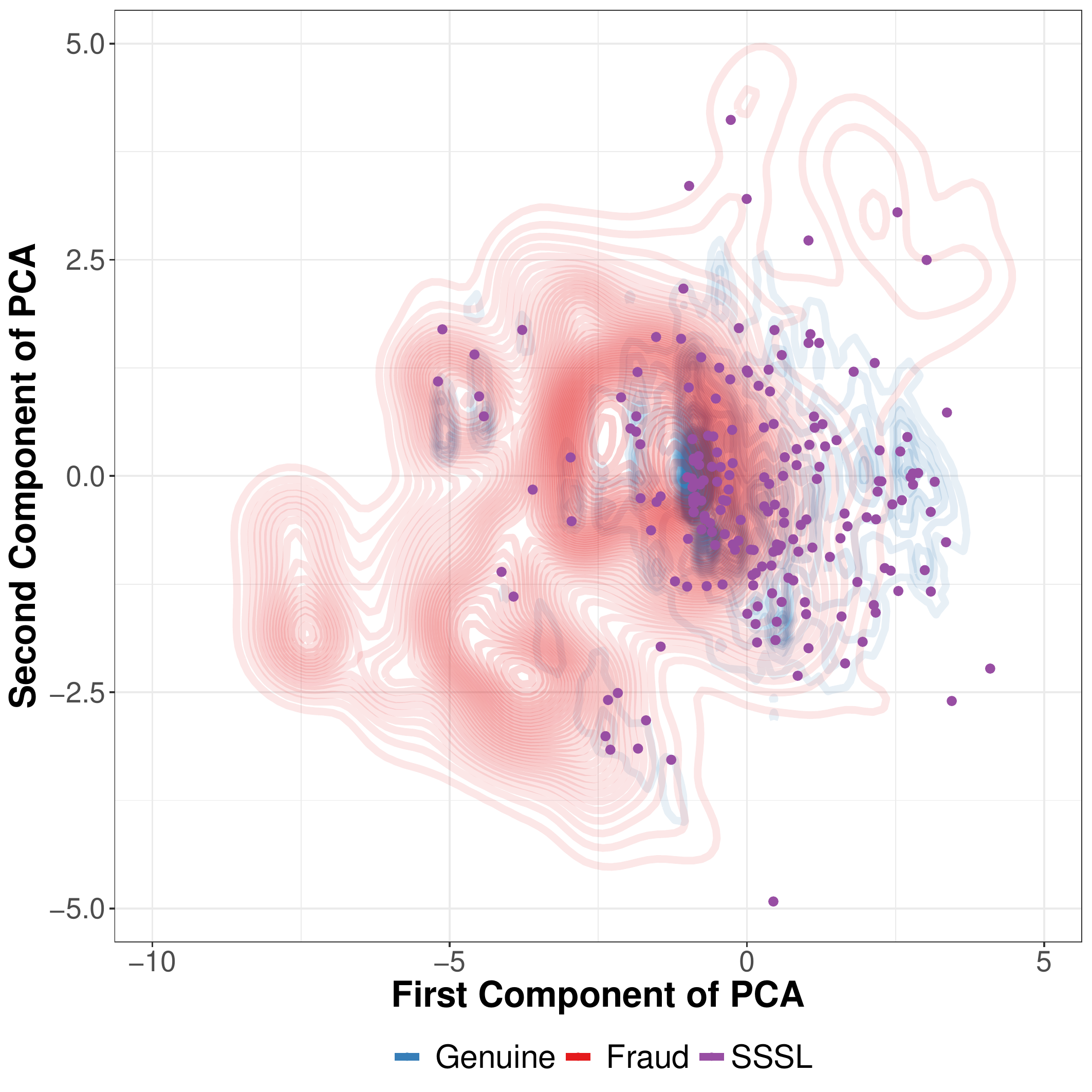}
\hfil
\caption{Class conditional distributions in PC1/PC2 space  and the set of 200 transactions selected using the Stochastic Semi-supervised Learning (SR) approach.
\label{fig:jpdf4}}
\end{figure}

On the basis of those considerations, it is possible to define a number of Stochastic Semi-Supervised strategies:
\begin{itemize}
\item SR:  no exploration budget ($q=0$) and attribution of the majority class to $m>0$ random transactions;
\item SU: no exploration budget ($q=0$) and  attribution of the majority class to the $m>0$ most uncertain points;
\item SM: no exploration budget ($q=0$) and attribution of the majority class to the $0.7 \times m$ most uncertain points and to $0.3 \times m$ random points;
\item SE: no exploration budget ($q=0$) and attribution of the majority class to the $m>0$ lowest risk points;
\end{itemize}

Additional variants can be created by simply allowing an exploration budget ($q>0$). The SR-U, SR-R and SR-M strategies are hybrid strategies which combine an exploration strategy (e.g. U in SR-U) and a SSSL strategy (e.g. SR in SR-U).

Fig.~\ref{fig:jpdf4} represents the set of 200 transactions selected by an SR approach. 
The transactions are selected randomly, and independently of the classifier $\CC$.
It is important to note that this sampling strategy basically follows the class conditional density of the Genuine class.
This is a consequence of the high imbalance of the dataset, since the distribution of frauds is negligible compared to the distribution of genuine transactions.

In \cite{jacobusse2016selection}, Jacobusse and Veenman presented additional variants of the \ac{SSSL}. 
They partition the original dataset ${\mathcal D}$ under study in three subsets:
\begin{itemize}
\item ${\mathcal D}^+$:  the subset of known positive samples, which corresponds in our scenario to the subset of frauds in the feedback set received by the investigators;
\item ${\mathcal D}^-$: the subset of known negative samples, which corresponds to the subset of genuine transactions in the feedback set received by the investigators;
\item ${\mathcal D}^0$: the subset of unlabeled samples, which corresponds to the set of unlabeled transactions.
\end{itemize}
While \ac{SSSL} proposes to add part of the samples in ${\mathcal D}^0$ and all the labeled samples in ${\mathcal D}^+$ and ${\mathcal D}^-$, the variant proposed by Jacobusse and Veenman suggests to remove ${\mathcal D}^-$ or part of it.
The idea is to reduce the selection bias which can arise when the process of selection is driven in an expert knowledge way.
In our context the expert knowledge selection is replaced by the classifier $\CC$.
In the experimental section, we will refer to this approach as SRN[$p$], with $p$ being the percentage of negative samples retained from the feedback set.

\subsection{Oversampling}
\label{subsection:os}

It is worth noting that a side-effect of the adoption of \ac{SSSL} (Section~\ref{subsection:sal}) is to add a number of majority class samples to the training set.
This goal is typically achieved by oversampling techniques, with the main difference that here the target class is the majority class and not the minority one.
In order to assess how SSSL situates with respect to conventional oversampling, we also consider a comparison with the two main oversampling techniques: Random Oversample (ROS)\cite{japkowicz2002class} and SMOTE \cite{chawlan2002smote}.
ROS consists in duplicating some random instances from the class to be oversampled until a given sample size is reached.
SMOTE creates artificial instances from the target class in the following manner: once the $k$ nearest neighbors from the same class have been identified, new artificial transactions are generated moving along the line segment joining the original instance and its $k$ neighbors.

\subsection{Multiple instance learning}
\label{subsection:mil}
This section deals with another specificity of the credit card fraud detection problem: the observations take place at the level of transactions but what is relevant for the company is the detection at the card level, since the investigation is performed at the card level and not at the transaction level.

From an AL perspective, since multiple transactions map to the same card, we could select query points by taking advantage of such one-to-many relationship.

\subsubsection{Querying by frequent uncertainty}
\label{subsection:qfu}

The rationale of \ac{QFU} boils down to query those cards which are mapped to the largest number of uncertain transactions.
We associate to each card $c$ a counter representing how many of its associated transactions $x_i \in c$ are uncertain, i.e. have a score $\probP_{\CC_{t}}(+|x_i) \in [0.5-v,0.5+v]$ where $v$ determines the size of the uncertainty range.
The counters are updated in real-time and the AL selection returns the $k$ cards with the highest counters.

\subsubsection{Combining function}
\label{subsection:mial}
A more advanced strategy to deal with card detection is inspired by~\cite{settles2008multiple}.
A \textit{combining function} can be used to aggregate all the posterior probabilities $p_i^c=\probP_{\CC}(+|x_{i})$ of the transactions $x_i \in c$ and derive the posterior probability $\probP_{\CC}(+|c)$.

The simplest combining function is the \emph{max} function (denoted MF), which returns
\begin{equation}
\label{maxp}
\probP_{\CC}^{MF}(+|c)= \max_{x_i \in c } p_i^c
\end{equation}


Alternatively, authors in  \cite{settles2008multiple} propose the \textit{softmax combining function}:
\begin{equation}
\label{SM}
\probP_{\CC}^{SM}(+|c)=\dfrac{\sum_{x_i} p^c_{i}e^{\alpha p^c_{i}}}{\sum_{x_i} e^{\alpha p^c_{i}}}
\end{equation}
where $\alpha$ is a constant that determines the extent to which \emph{softmax} approximates a \emph{max} function.

In order to i) increase the sensitivity of the card risk to high risk transactions and to ii) reduce its sensitivity to low risk  transactions, we propose a \textit{logarithmic combining function} returning the score
\begin{equation}
\label{CF}
 \sum_{x_i \in c }
  -\dfrac{1}{\log{s^c_{i}}}
\end{equation}
where 
$s^c_{i}=\begin{cases}
    p^c_{i} - \epsilon   & \quad \text{if } p^c_{i} > 0.5\\
    \epsilon & \quad \text{otherwise } \\
  \end{cases}
$
and $\epsilon$ is a very small number.

\begin{table}[!h]
\fontsize{9}{9}\selectfont
\setlength{\tabcolsep}{1em}
\renewcommand{\arraystretch}{1.3}
\caption{Scoring of transactions}
\label{ranktrx1}
\centering
\begin{tabular}{|c|c|c|c|}
\hline
\textbf{Rank}& \textbf{Card Id} & \textbf{Trx Id} & \textbf{$p^c_{i}$} \\
\hline
1&A & A7 & 0.90 \\
2&B & B3 & 0.88 \\
3&B & B5 & 0.87 \\
4&A & A2 & 0.83 \\
... & ... & ...& ... \\
\hline
\end{tabular}
\end{table}

\begin{table}[!h]
\fontsize{9}{9}\selectfont
\setlength{\tabcolsep}{1em}
\renewcommand{\arraystretch}{1.3}
\caption{Scoring  of cards on the basis of transactions from Table~\ref{ranktrx1} with three combining functions}
\label{combined1}
\centering
\begin{tabular}{|c|c|c|c|}
\hline
 \textbf{Card} & \textbf{max($p^c_{i}$)} & \textbf{softmax($p^c_{i}$)} & \textbf{log. ($p^c_{i}$)} \\
\hline
A & \textbf{0.90} & 0.87 & 34.21\\
B & 0.88 & \textbf{0.88} & \textbf{34.55}\\
... & ... & ...& ... \\
\hline
\end{tabular}
\end{table}

Table \ref{combined1} illustrates the scores associated to the transactions of Table \ref{ranktrx1} for the three combining functions presented above.
It appears that, unlike the \emph{max} function, the other two functions are able to take into account the impact of multiple risky transactions
on the overall risk of a card. In other terms two high risk transactions weight more than a simple one with a marginal higher risk.
However, the softmax and the logarithmic functions differ in the importance they give to low risk transactions.
Suppose we add a low risk transaction (Table \ref{ranktrx2}) for card ``B'' to the set of transactions of Table \ref{ranktrx1}.
Table~\ref{combined2} shows that the sensitivity of the card risk to such additional transaction is much larger in the \emph{softmax} than in the logarithmic case.
The counter-intuitive consequence is that according to the \emph{softmax} function the card ``B'' becomes now less risky than the card ``A''.

\begin{table}[!h]
\fontsize{9}{9}\selectfont
\setlength{\tabcolsep}{1em}
\renewcommand{\arraystretch}{1.3}
\caption{Additional transaction}
\label{ranktrx2}
\centering
\begin{tabular}{|c|c|c|c|}
\hline
\textbf{Rank}& \textbf{Card Id} & \textbf{Trx Id} & \textbf{$p^c_{i}$} \\
\hline
20000&B & B6 & 0.20 \\
\hline
\end{tabular}
\end{table}

\begin{table}[!h]
\fontsize{9}{9}\selectfont
\setlength{\tabcolsep}{1em}
\renewcommand{\arraystretch}{1.3}
\caption{Scoring  of cards on the basis of transaction from Tables~\ref{ranktrx1} and~\ref{ranktrx2} with three combining functions}
\label{combined2}
\centering
\begin{tabular}{|c|c|c|c|}
\hline
 \textbf{Card} & \textbf{max($p^c_{i}$)} & \textbf{softmax($p^c_{i}$)} & \textbf{log. ($p^c_{i}$)} \\
\hline
A & \textbf{0.90} & \textbf{0.87} & 34.21\\
B & 0.88 & 0.74 & \textbf{34.55}\\
... & ... & ...& ... \\
\hline
\end{tabular}
\end{table}

\section{Experiments}
\label{section:exp}

\begin{figure*}[!ht]
\label{fig:outl}
\centering
\subfloat[]{\includegraphics[width=3in, height=3.1in]{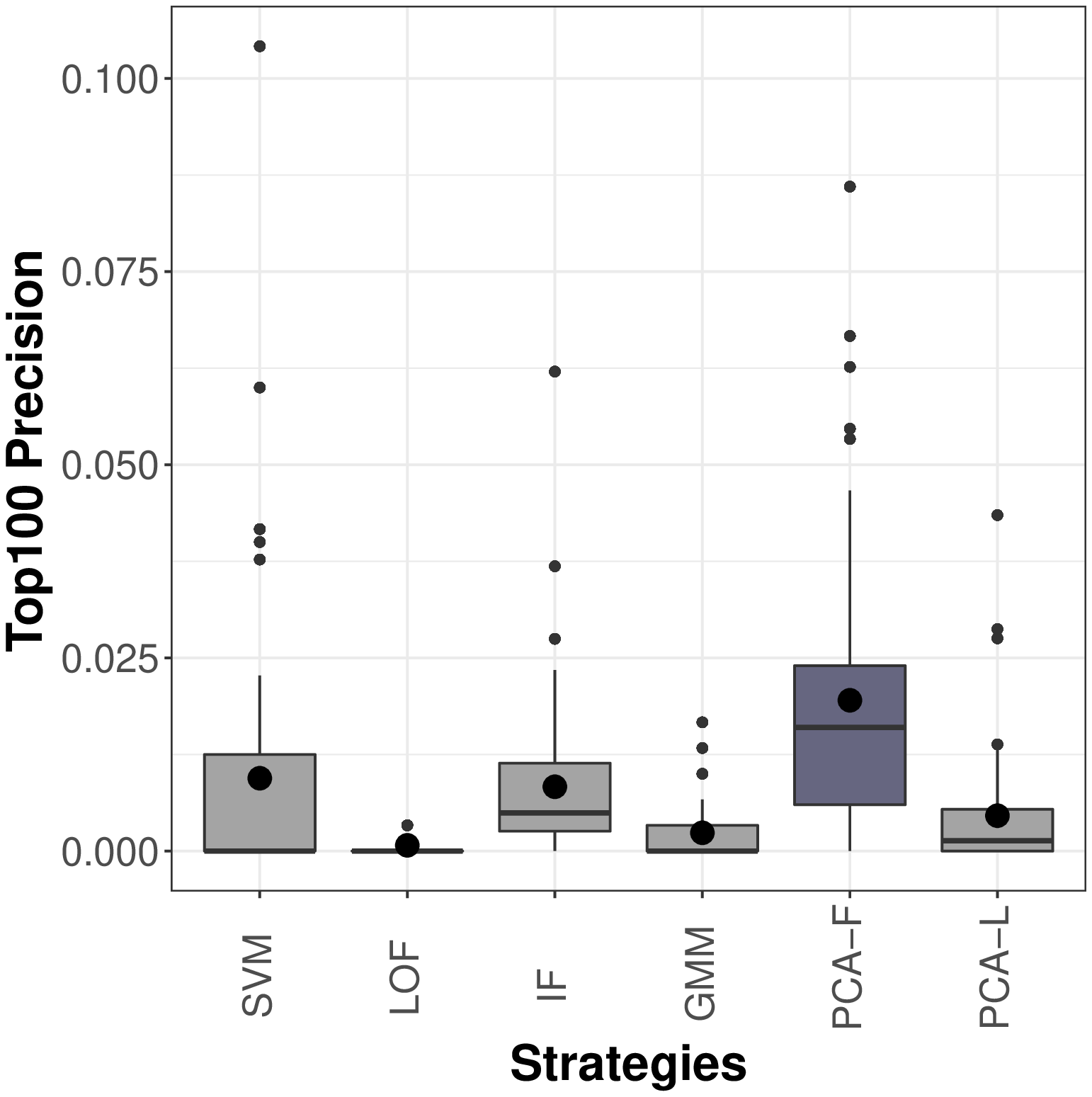}
\label{fig:outlprec}}
\subfloat[]{\includegraphics[width=1.9in, height=3.1in]{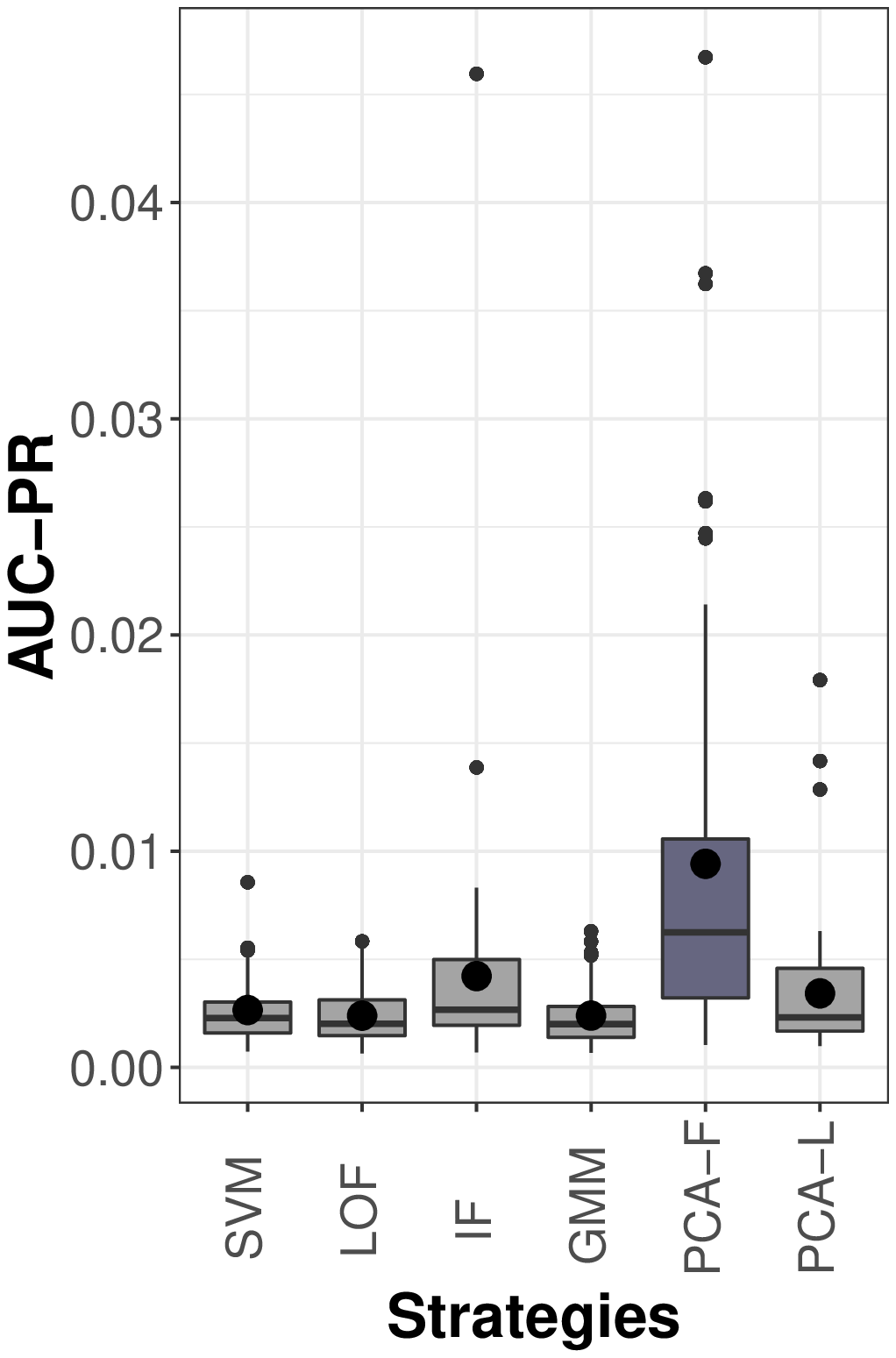}
\label{fig:outlpr}}
\subfloat[]{\includegraphics[width=1.9in, height=3.1in]{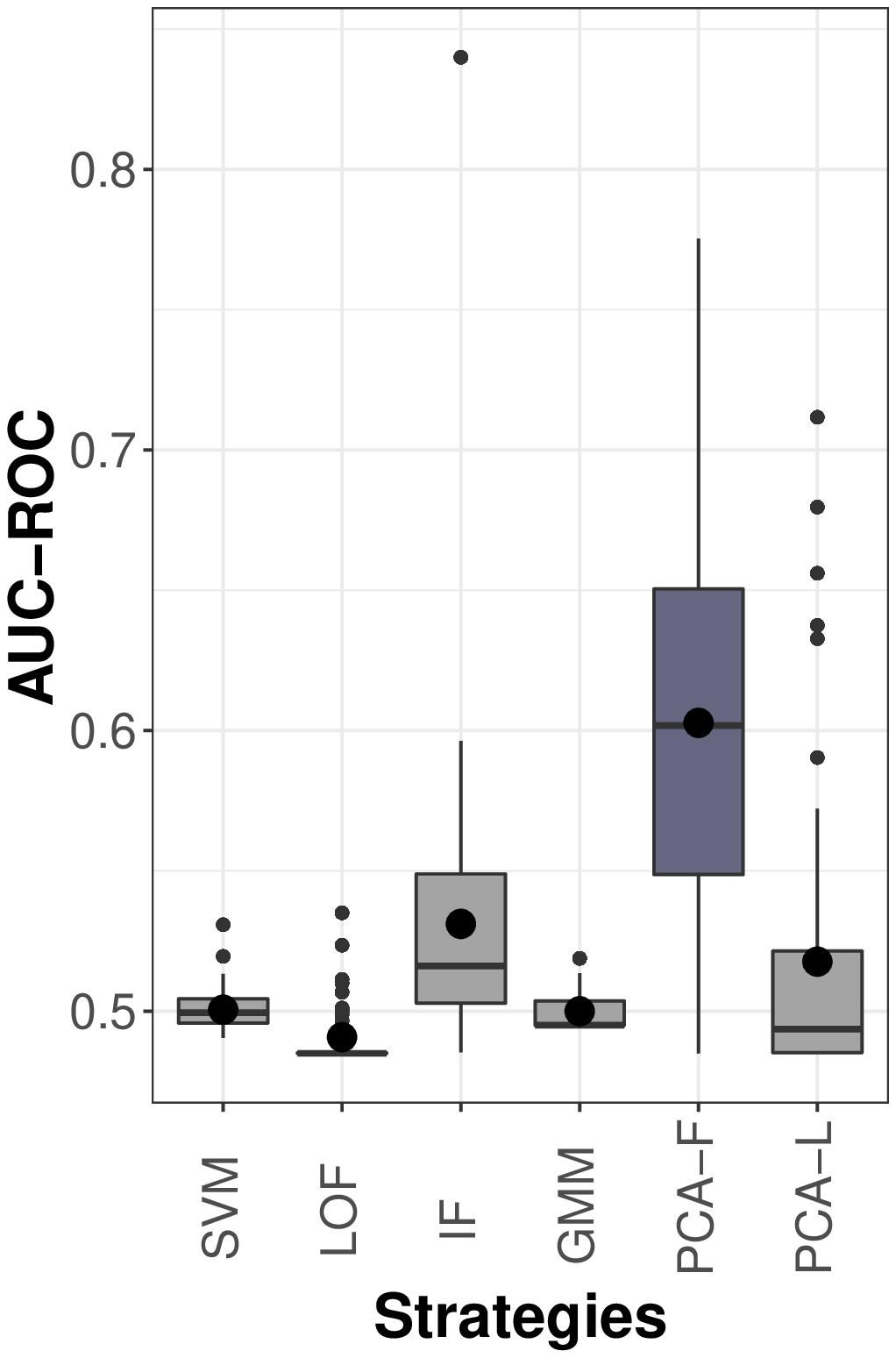}
\label{fig:outlroc}}
\hfil
\caption{Transaction-based case study with unsupervised passive learning. Box-plots summarize the accuracy measures obtained over 60 days and 20 trials. Black points indicate the mean value for each box. Comparison in terms of: Top100 Precision (a), Area Under the Precision-Recall Curve (b) and Area Under the Receiver Operator Characteristic Curve (c). Dark boxes indicate the best strategy (paired Wilcoxon test).}\label{fig:outl}
\end{figure*}

This section relies on a large imbalanced dataset of 12 million credit card transactions provided by our industrial partner Worldline.

In this realistic case-study, only a very small number ($k=100$) of cards per day can be queried, amounting to roughly 0.2\% of labeled points.
The dataset has 32 features and it covers 60 days, each day including roughly $200K$  transactions.

Two sets of experiments are performed: the first measures the detection accuracy at the level of the transactions, while the second measures the detection accuracy at the card level. In the first study, for the sake of simplicity, the classification model $\CC$ is a conventional random forest model $\RF$ while a more realistic model $\AWC$ (discussed in \cite{dal2017TNNLS} and in Section \ref{section:ps})
is used for the cards\footnote{The use of two different learning strategies is justified by the need to assess the robustness of the AL strategies with respect
to different learning methods and different detection tasks (transaction-based and card-based). }.
Since the randomization process in $\RF$ and $\AWC$ may induce variability in the accuracy assessment, we present the results of twenty  repetitions of the streaming.

All the AL strategies are compared in identical situations and initialized with the same random and balanced set (initial training set $D$ presented in algorithm \ref{alg:process}).
The results are presented as box-plots summarizing the fraud detection performance over the 60 days.
In particular we have considered the following accuracy measures:  Top100 Precision,  Area Under the Precision-Recall Curve (AUC-PR) and  Area Under the Receiver Operator Characteristic Curve (AUC-ROC).
In all the plots, the dark boxes are used to denote the most accurate AL strategy as well as the ones which do not differ significantly from it (paired Wilcoxon signed rank test with 5\% significance level).

The precision over the Top100 alerts is expected to be larger for $\RF$ than $\AWC$ since multiple positive alerts for the same card will be accounted as several true positives in the transaction case but as a single success in the card case. 
We made all the code available on Github\footnote{https://github.com/fabriziocarcillo/

StreamingActiveLearningStrategies}.

\begin{figure*}[!ht]
\label{fig:trx}
\centering
\subfloat[]{\includegraphics[width=3in, height=3.1in]{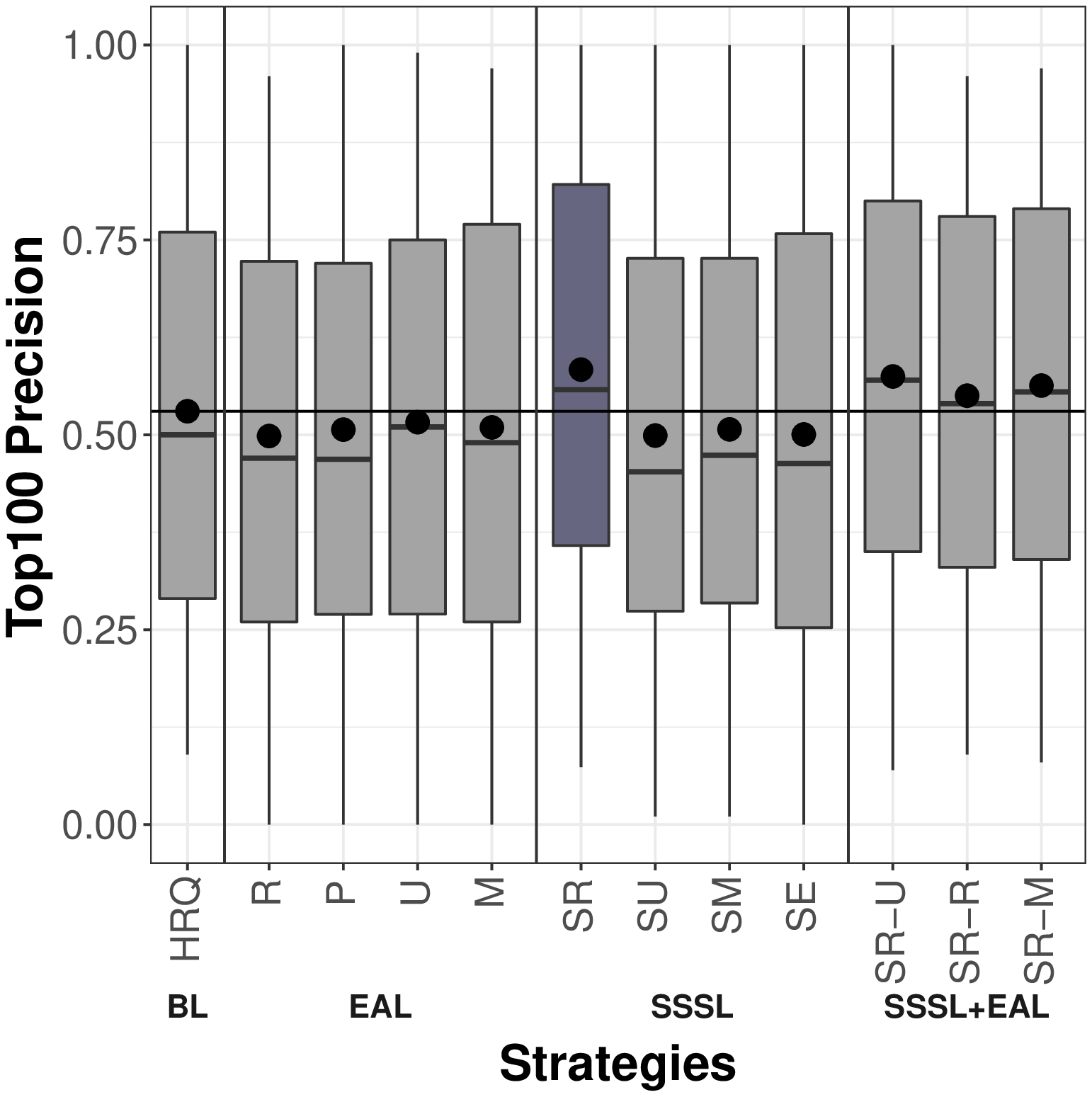}
\label{fig:trxprec}}
\subfloat[]{\includegraphics[width=1.9in, height=3.1in]{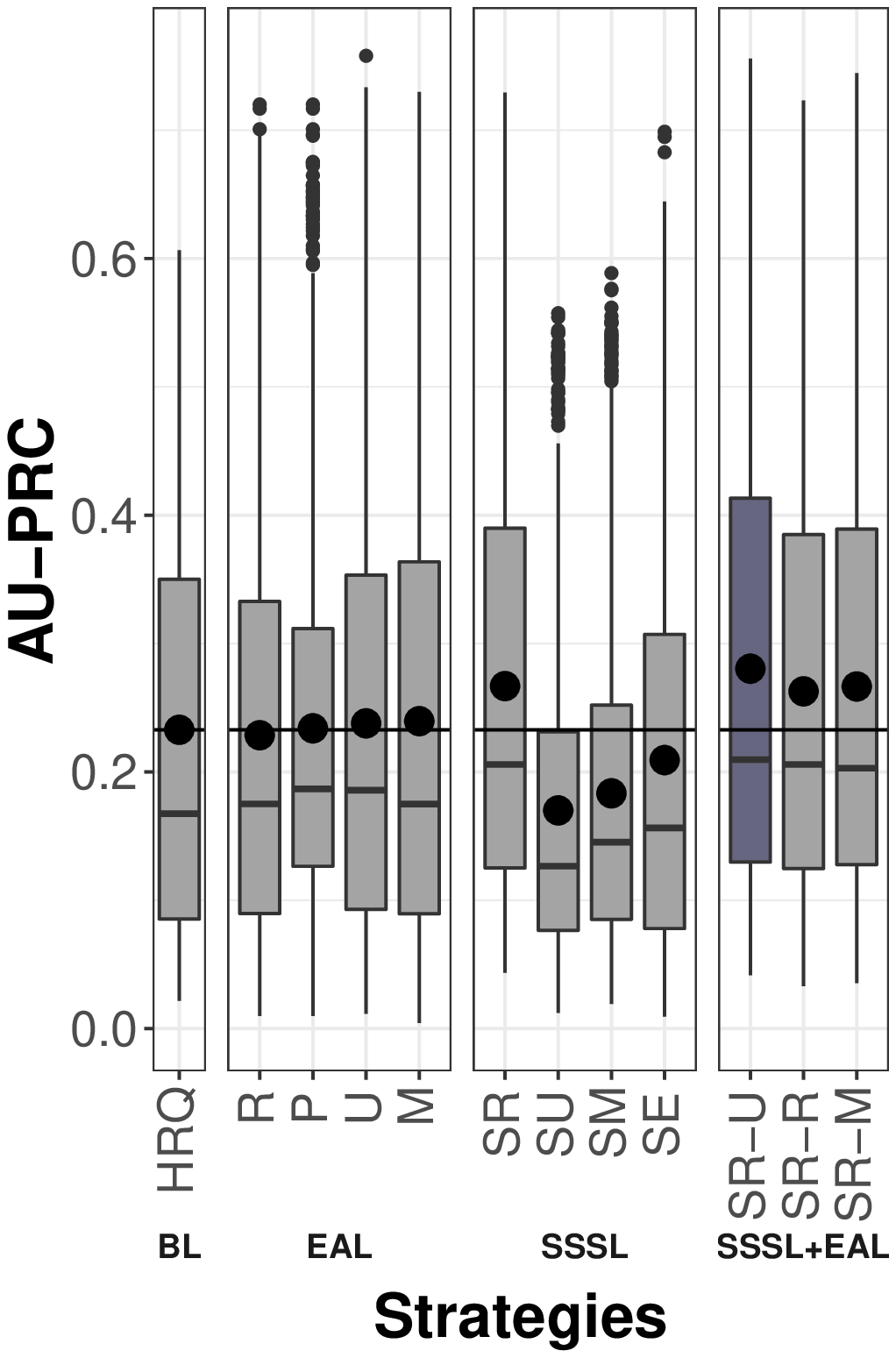}
\label{fig:tpr}}
\subfloat[]{\includegraphics[width=1.9in, height=3.1in]{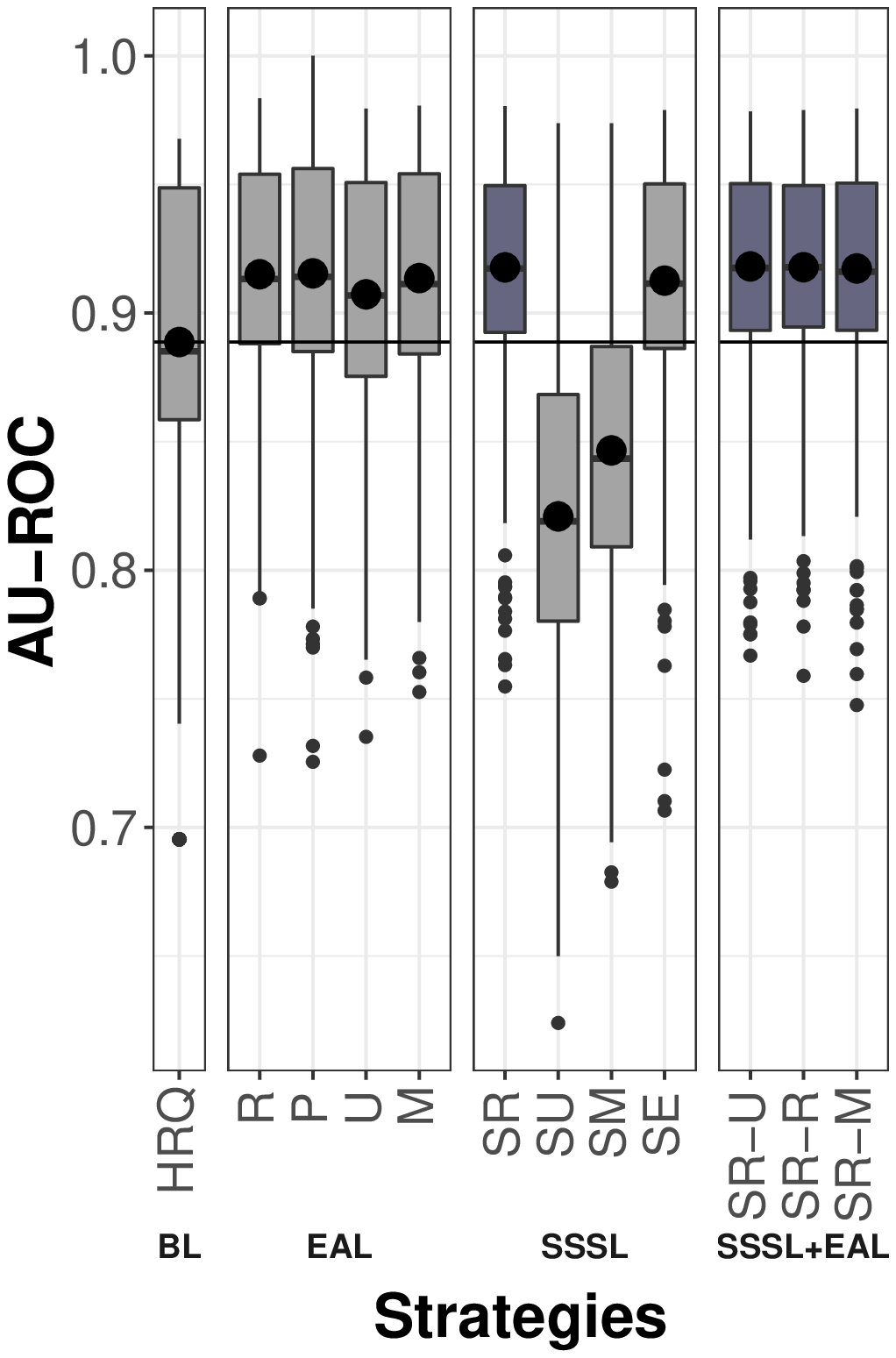}
\label{fig:troc}}
\hfil
\caption{Transaction-based case study. Box-plots summarize the accuracy measures obtained over 60 days and 20 trials. Black points indicate the mean value for each box and the horizontal line indicates the mean for the baseline \ac{HRQ}. The extended names for the strategies listed on the horizontal axes can be found in Table \ref{syntstr}. Comparison in terms of: Top100 Precision (a), Area Under the Precision-Recall Curve (b) and Area Under the Receiver Operator Characteristic Curve (c). 
Dark boxes indicate the best strategy and those which are not statistically different (paired Wilcoxon test).
\label{fig:trxall}}
\end{figure*}

\begin{figure*}[!t]

\centering
\subfloat[]{\includegraphics[width=3in, height=3.1in]{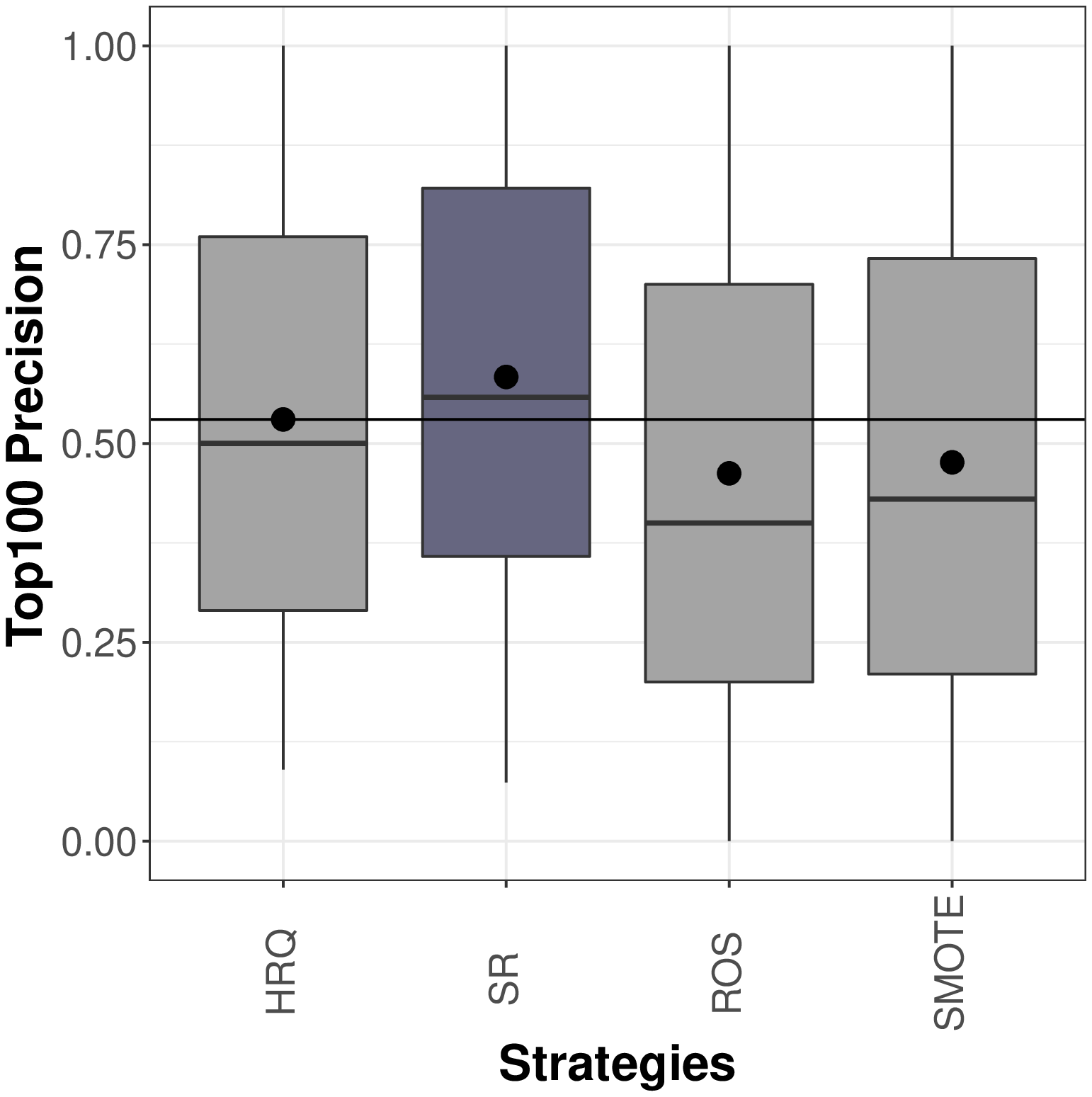}
\label{fig:smoteprec}}
\subfloat[]{\includegraphics[width=1.9in, height=3.1in]{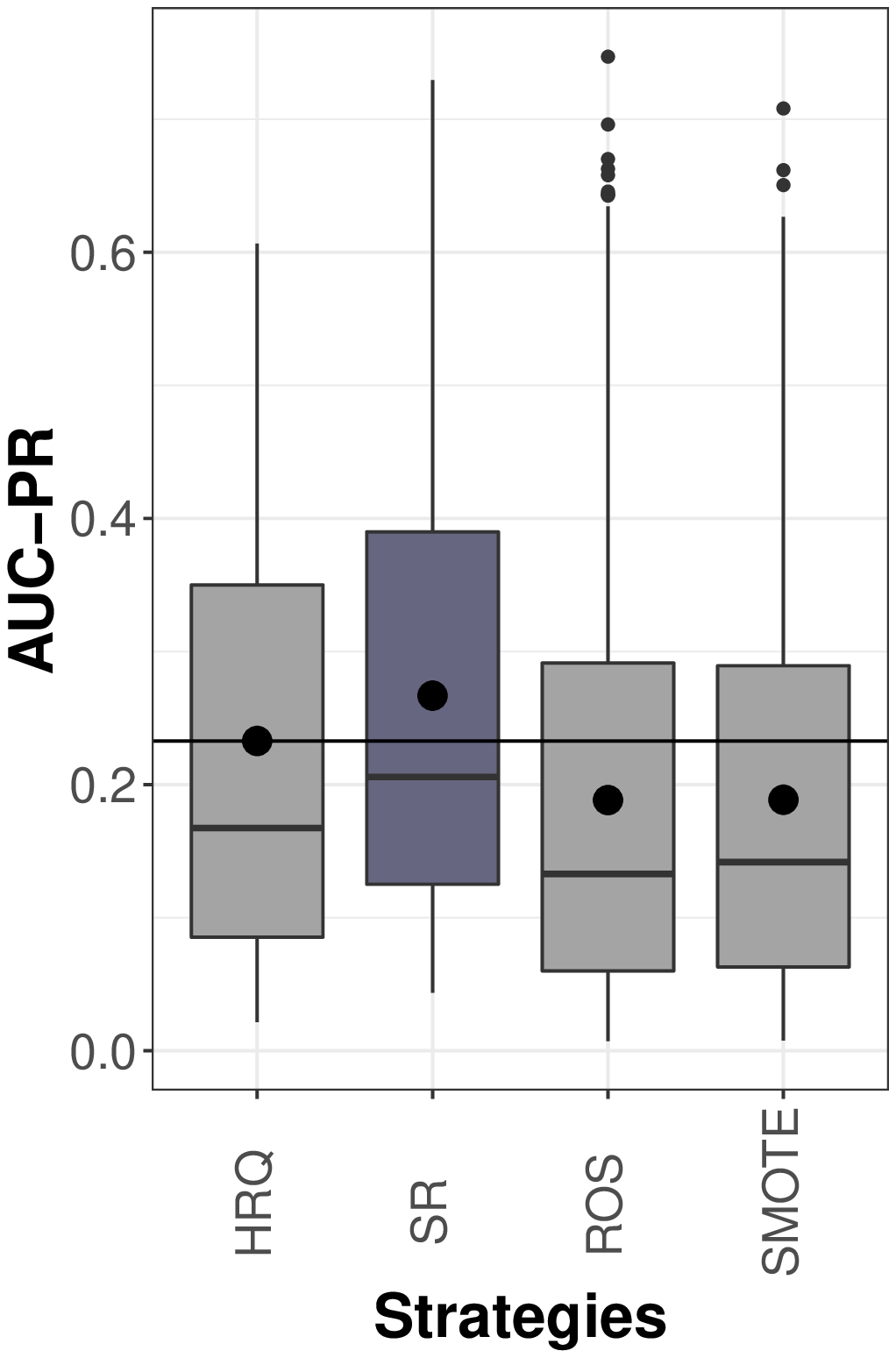}
\label{fig:smotepr}}
\subfloat[]{\includegraphics[width=1.9in, height=3.1in]{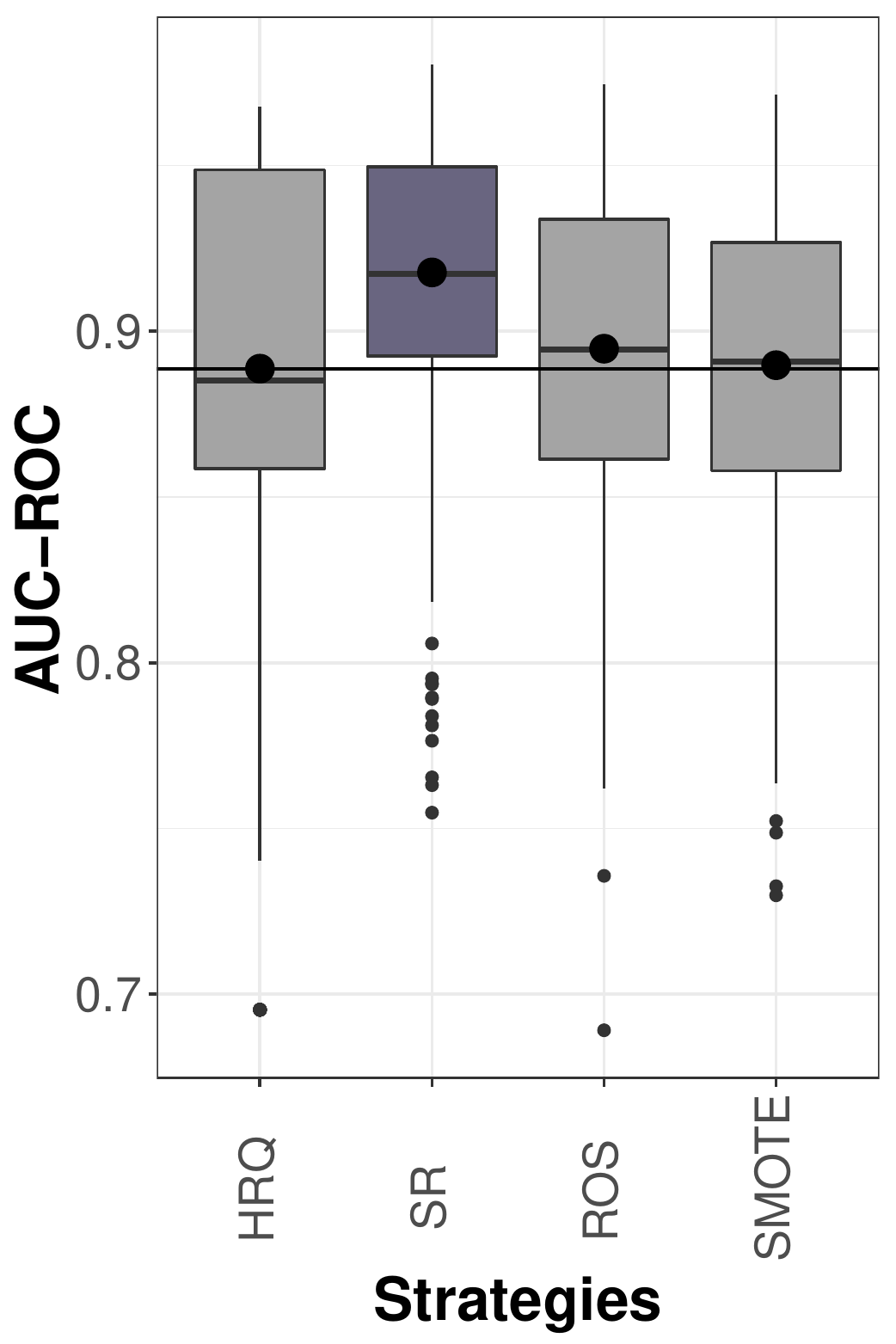}
\label{fig:smoteroc}}
\hfil
\caption{Transaction-based case study. Comparison in terms of: Top100 Precision (a), Area Under the Precision-Recall Curve (b) and Area Under the Receiver Operator Characteristic Curve (c). Dark boxes indicate the best strategy and those which are not statistically different (paired Wilcoxon test).\label{fig:smote}}

\end{figure*}

\subsection{Transaction-based fraud detection}
\label{subsection:tfd}

In order to better situate the results of AL techniques, we start by reporting the accuracy of several passive (Section \ref{subsection:outdet}) unsupervised methods. 
In Fig.~\ref{fig:outl}, we show the accuracy of six state-of-the-art unsupervised outlier detection techniques: One-class Support Vector Machine (SVM), Local Outlier Factor (LOF), Isolation Forest (IF), Gaussian Mixture Models (GMM), First Component of the PCA (PCA-F) and Last Component of the PCA (PCA-L).
Though PCA-F works better in terms of Top100 Precision, AUC-PR and AUC-ROC, overall the accuracy of passive unsupervised learning is very low.

In Fig.~\ref{fig:trxall}, we report the fraud detection accuracy of the AL techniques discussed in Section~\ref{section:str}. 
A horizontal line is added in order to make the comparison with the baseline strategy HRQ easier.
The experiments are performed with $k=100$, $q=5$ and $m=1000$. These hyper-parameters
have been set by trial-and-error and are compatible with the kind of exploration effort that our industrial partner could ask to its investigators.

\begin{figure}[!h]
\centering
\includegraphics[width=3.2in, height=3.2in]{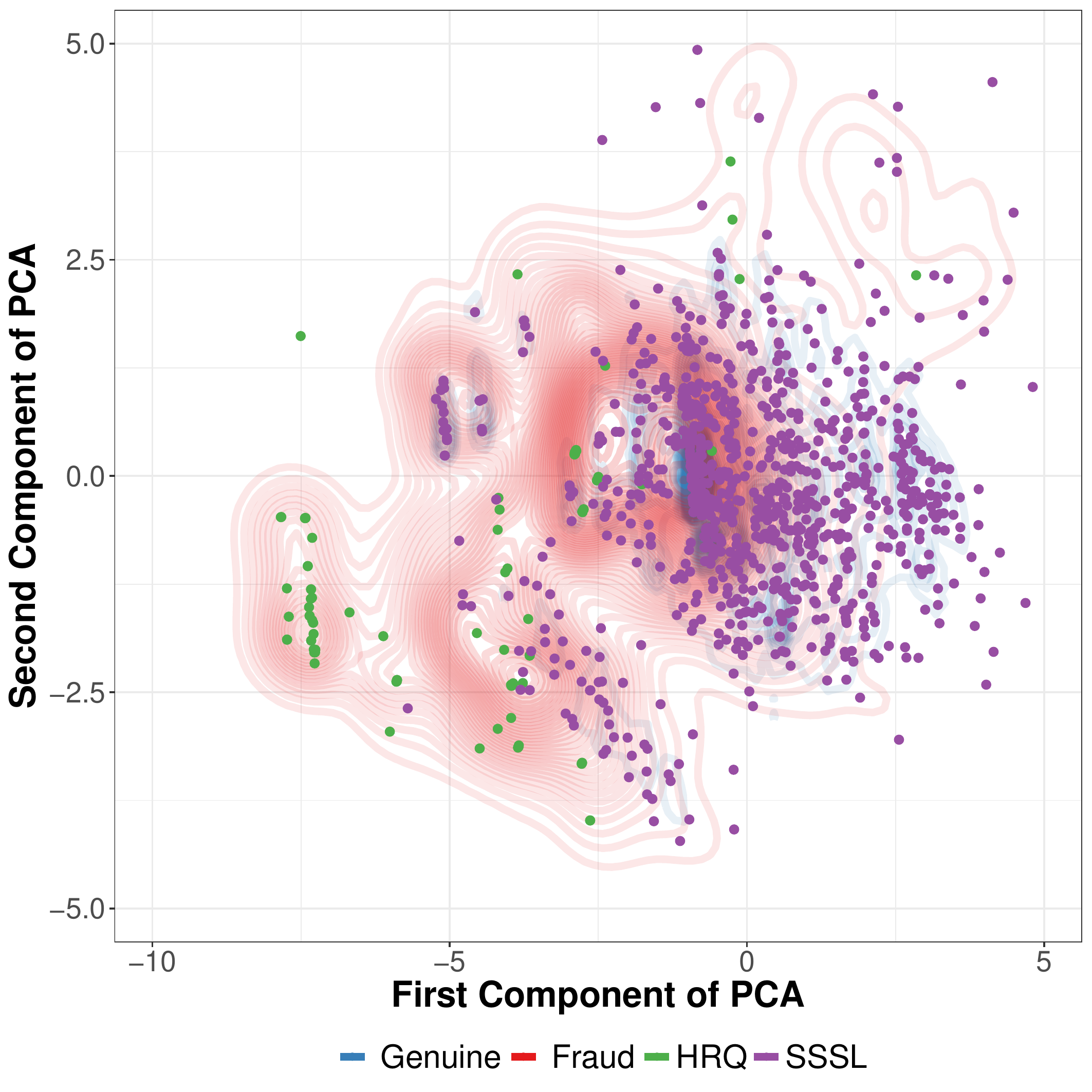}
\hfil
\caption{Class conditional distributions in the PC1/PC2 space and the transactions selected by  the SR strategy, the best one in our experimental comparison. A set of 100 transactions is selected using the Highest Risk Querying approach (green dots), and a random set of 1000 transactions is selected and labeled as the genuine class (SSSL, purple dots).
\label{fig:jpdf6}}
\end{figure}

The first remark is that AL (Fig.~\ref{fig:trxall}) significantly outperforms passive learning unsupervised techniques (Fig.~\ref{fig:outl}).
Second, it appears that exploratory AL alone is not able to outperform the standard \ac{HRQ} strategy.
In particular unsupervised explorations are of no use: the PCA based exploration EAL-P as well as random exploration EAR are significantly worse than the baseline HRQ.
The highest accuracy is obtained by \ac{SSSL} (SR) or by combining SSSL with uncertainty sampling (SR-U).
The SR strategy leads to an improvement in precision of 5.84\%, while SR-U leads to an improvement of 5.15\%.
Similar improvement are observed for the AUC-PR (Fig.~\ref{fig:tpr}), while a wider range of techniques perform better in terms of AUC-ROC curve (Fig.~ \ref{fig:troc}).
The most efficient combination in our setting is therefore obtained by a combination of stochastic semi-supervised approach with the standard \ac{HRQ} strategy for active learning (Fig.~\ref{fig:jpdf6}).

We have also compared the semi-supervised technique SR with the standard Random Oversample and SMOTE oversampling techniques.
As shown in Fig.~\ref{fig:smote}, SR appears to be better in terms of Top100 Precision, AUC-PR and AUC-ROC.
ROS and SMOTE outperform \ac{HRQ} only in terms of AUC-ROC.

\begin{figure*}[!ht]
\centering
\subfloat[]{\includegraphics[width=3in, height=3.1in]{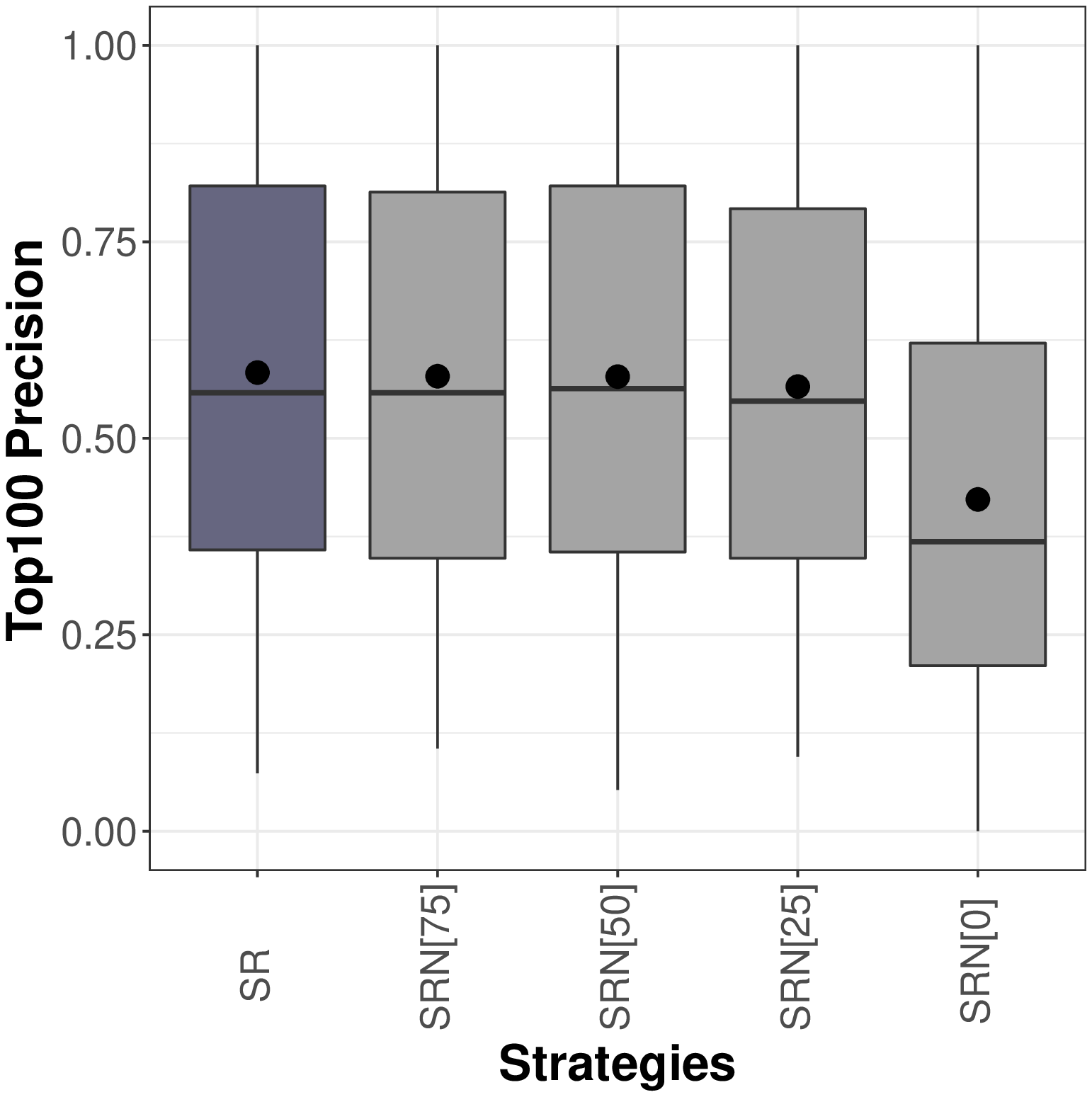}
\label{fig:srnprec}}
\subfloat[]{\includegraphics[width=1.9in, height=3.1in]{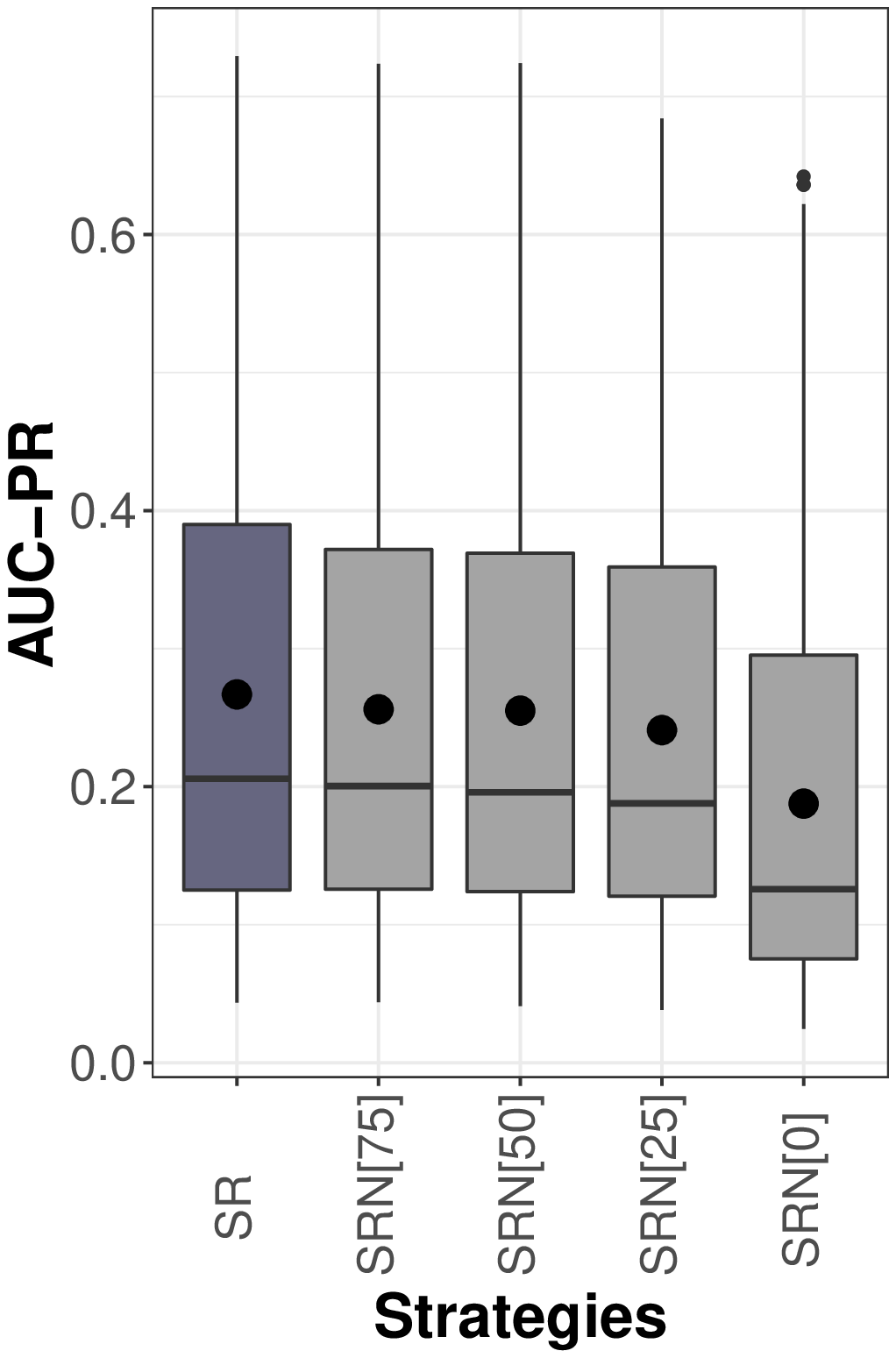}
\label{fig:srnpr}}
\subfloat[]{\includegraphics[width=1.9in, height=3.1in]{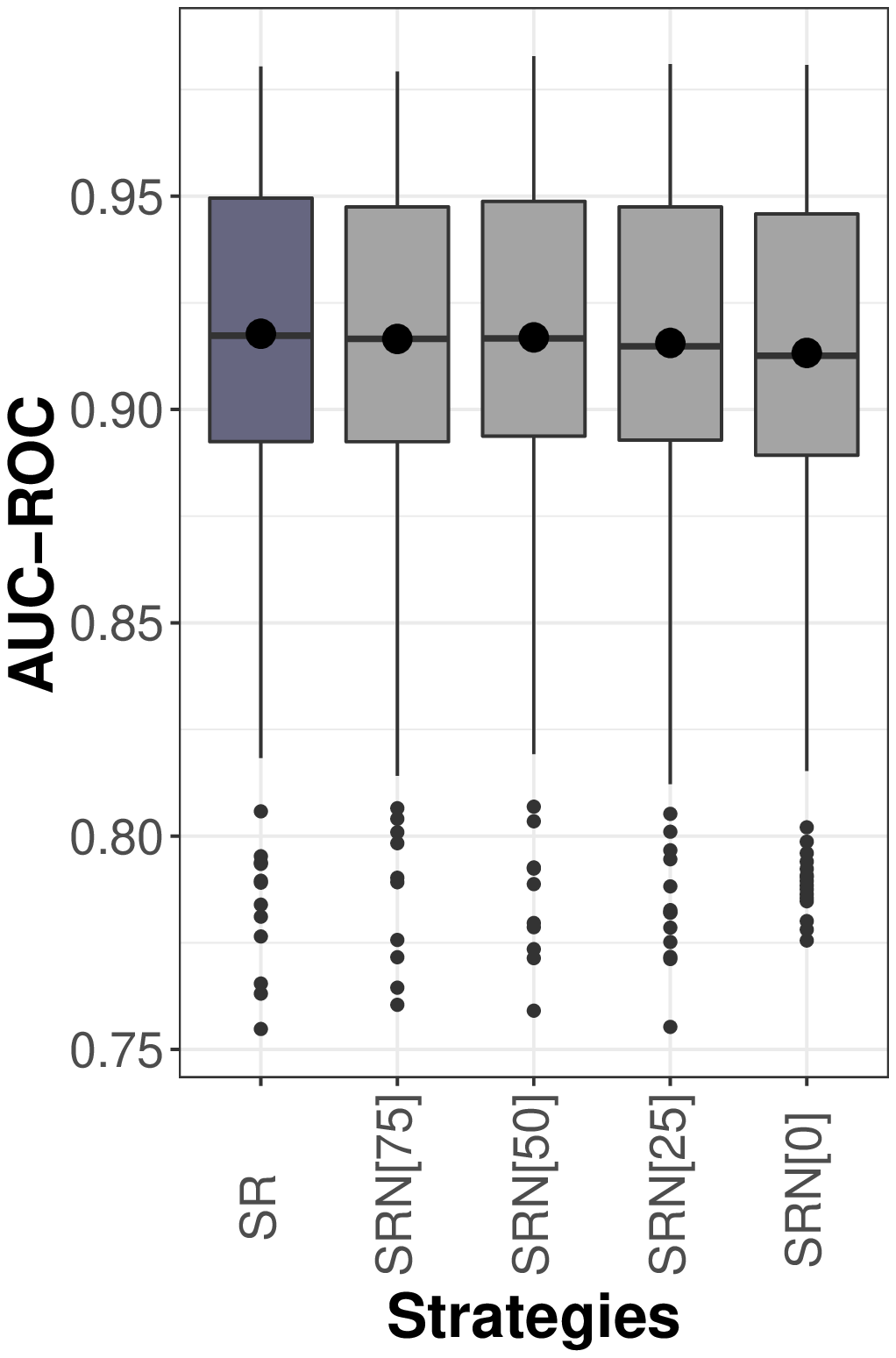}
\label{fig:srnroc}}
\hfil
\caption{Transaction-based case study. Comparison of sampling strategies SRN[$p$] for different $p$ values: percentage of retained negative samples from the feedback set. Note that SR corresponds to SRN[$p$] with $p$=100. Comparison in terms of: Top100 Precision (a), Area Under the Precision-Recall Curve (b) and Area Under the Receiver Operator Characteristic Curve (c). Dark boxes indicate the best strategy and those which are not statistically different (paired Wilcoxon test).\label{fig:srn}}

\end{figure*}

In Fig.~\ref{fig:srn}, a comparison between SR and SRN[$p$] with different values of $p$ is reported (where $p$ is the percentage of genuine transactions retained from the feedback set).
SR is statistically better than the SRN[$p$] strategies in terms of Precision, AUC-PR and AUC-ROC and for all $p \in \{75, 50, 25, 0\}$.
Our findings are similar to the results of Jacobusse and Veenman \cite{jacobusse2016selection}.
They have used a synthetic dataset with an imbalance ratio of 1\% and they have observed a higher AUC-ROC for the SRN[$p$] strategy in case of very small $p$.

\begin{figure}[!h]
\centering
\includegraphics[width=3in, height=3in]{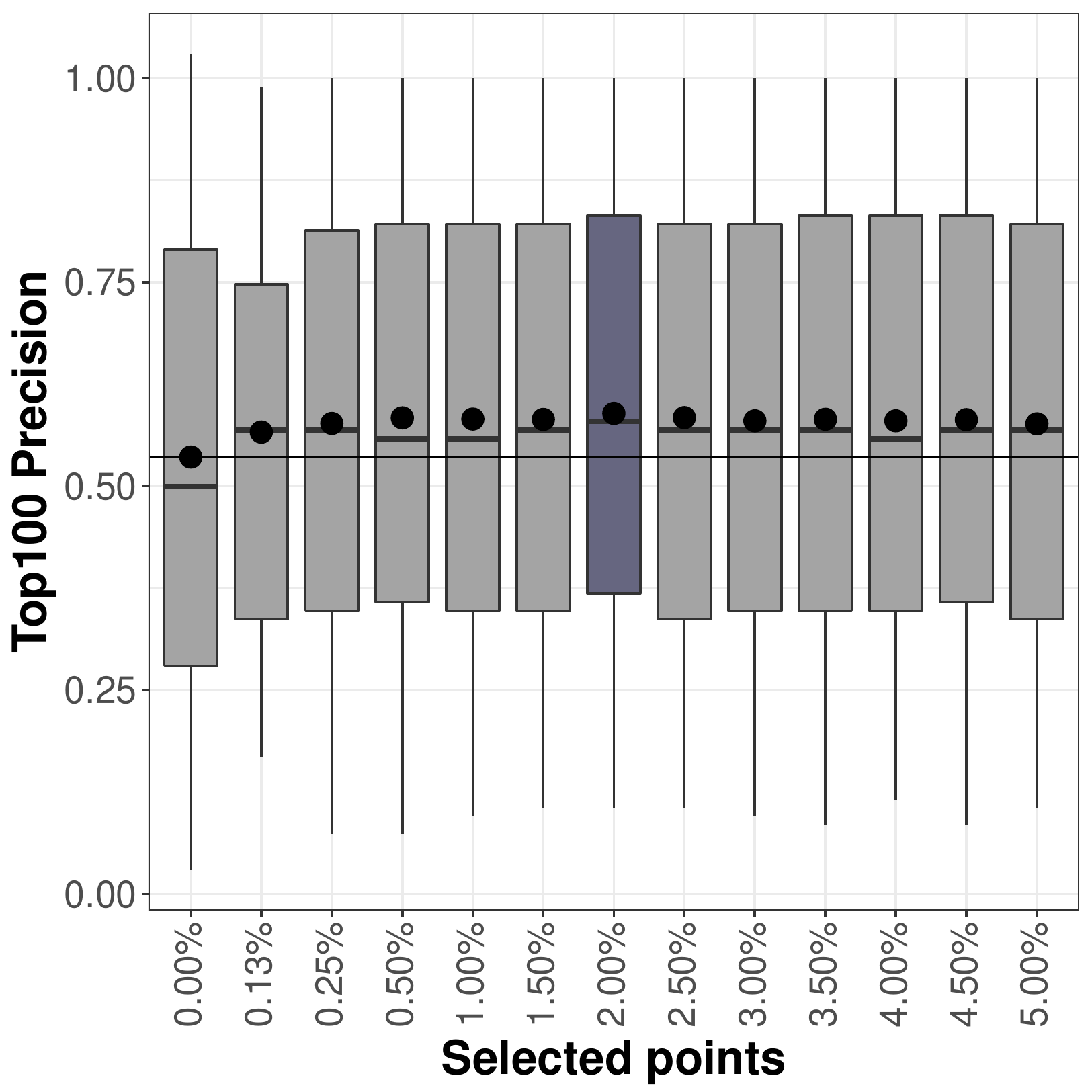}
\hfil
\caption{Transaction-based case study. Top100 Precision when varying the number of points selected in SR. The dark box indicates the best strategy (paired Wilcoxon test).\label{fig:nptrx}}
\end{figure}

In Fig.~\ref{fig:nptrx}, we report the impact on precision of the proportion of daily transactions selected by SR.
The SR experiments reported in Fig.~\ref{fig:trxall} refer to a selection of 1000 points
(0.50\% of the daily transactions) while the highest precision (statistically significant) is attained for 2\% of the daily transactions.
The figure shows then the existence of a trade-off in precision between a  configuration corresponding to  HRQ (i.e. no additional selected points, on the left-side)   and a configuration with many  added points  (right side).
Our interpretation is that both selection bias (left-side configuration) and excessive imbalance (right-side configuration) may be detrimental to precision. An accurate tuning of such trade-off is expected to be beneficial for the final performance.

\begin{figure*}[!t]
\centering
\subfloat[]{\includegraphics[width=3in, height=3.1in]{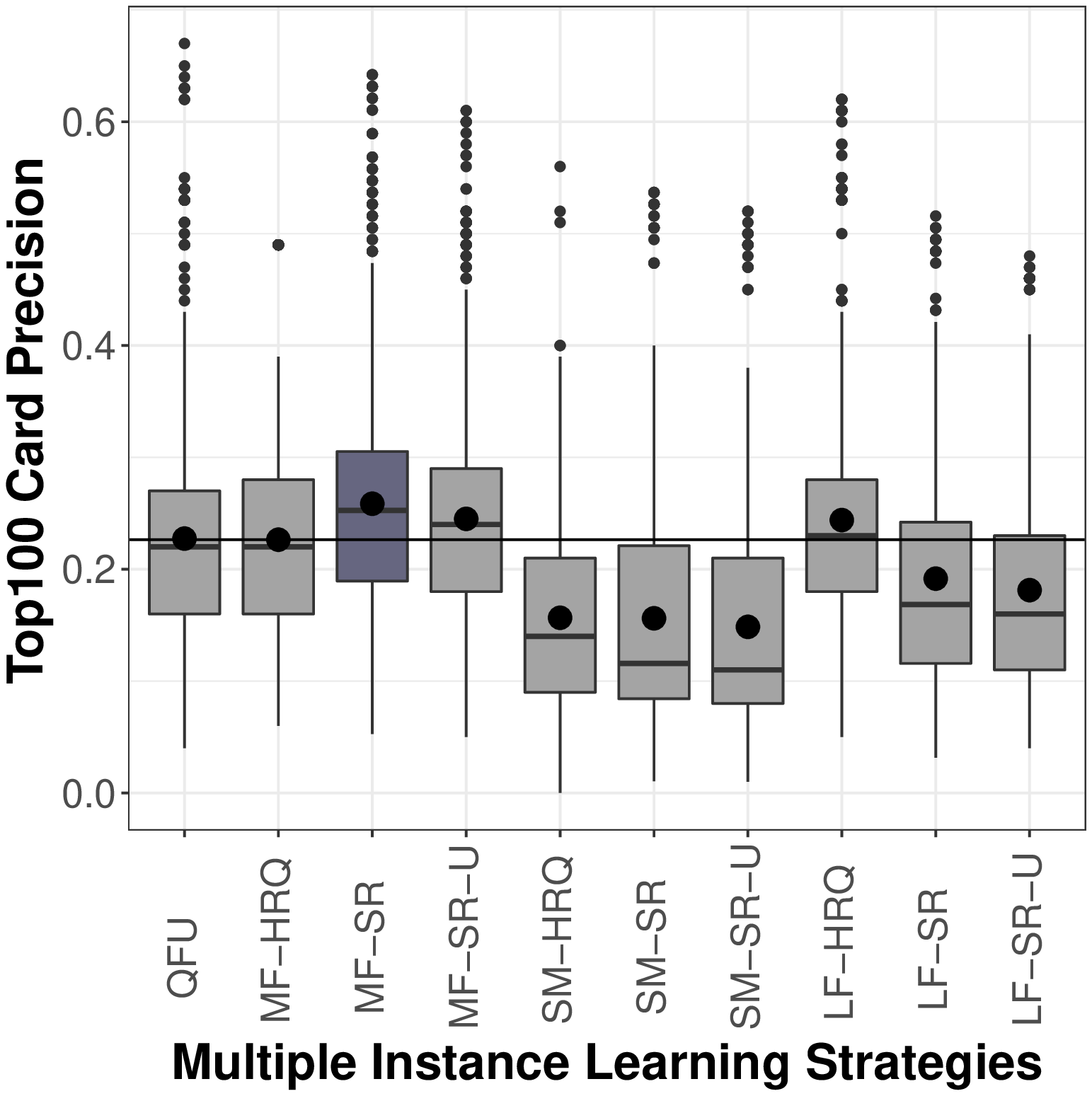}
\label{fig:cprec}}
\subfloat[]{\includegraphics[width=1.9in, height=3.1in]{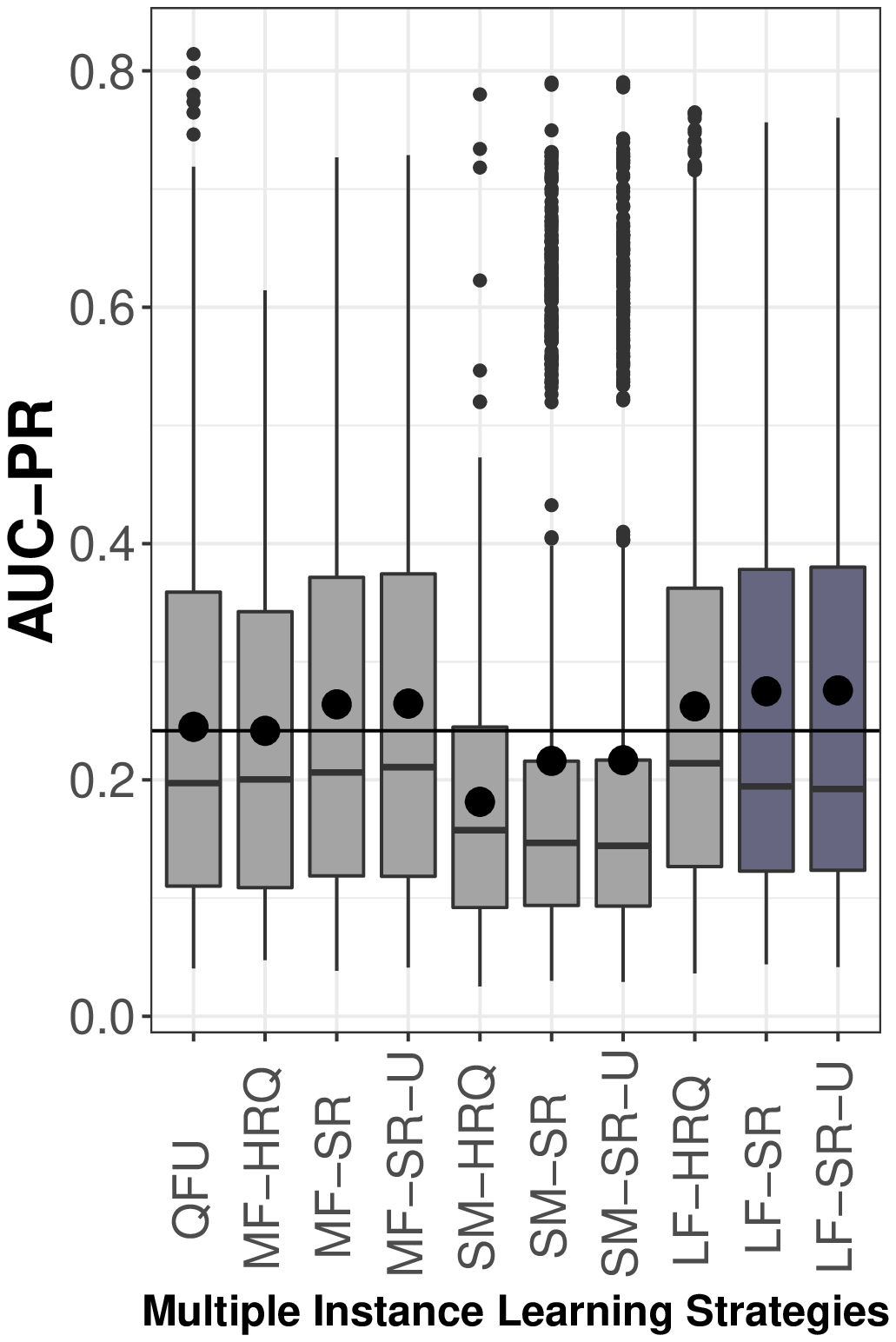}
\label{fig:cpr}}
\subfloat[]{\includegraphics[width=1.9in, height=3.1in]{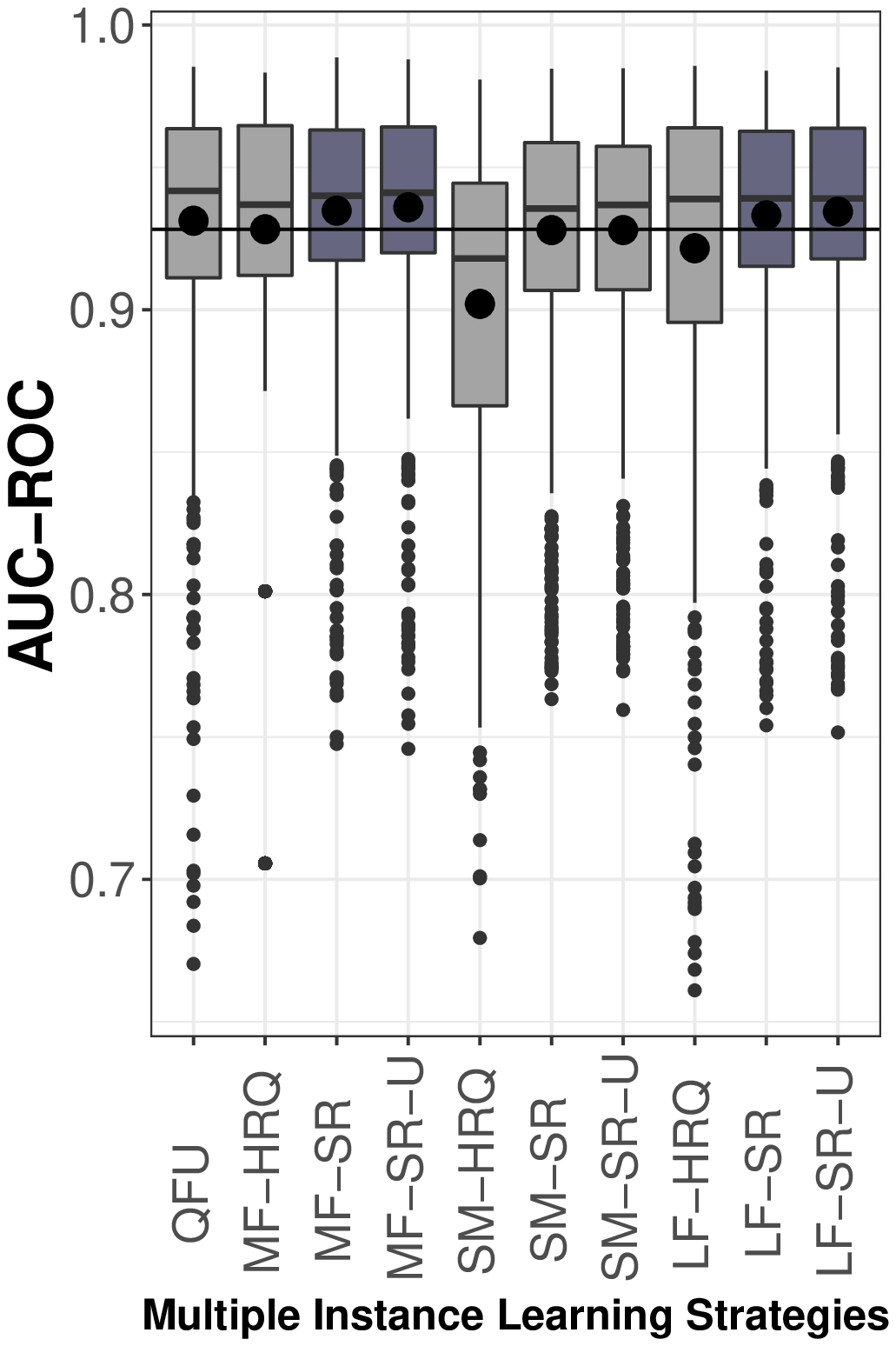}
\label{fig:croc}}
\hfil
\caption{Card-based case study. Comparison in terms of: Top100 Precision (a), Area Under the Precision-Recall Curve (b) and Area Under the Receiver Operator Characteristic Curve (c). Dark boxes indicate the best strategy and those which are not statistically different (paired Wilcoxon test).\label{fig:crdall}}
\end{figure*}

\subsection{Card-based fraud detection}
\label{subsection:tfd}
In this second case study, we retain the most promising techniques form the transaction assessment (namely, SR and SR-U), and we
compare them with the Multiple Instance Learning strategies described in Section~\ref{subsection:qfu}  ($v=0.05$) and~\ref{subsection:mial} ($\epsilon=1e-3$).
Fig.~\ref{fig:crdall} summarizes the results for the card detection study, using Top100 Card Precision, AUC-PR and AUC-ROC as accuracy metrics. 


When the Top100 Precision is considered as performance metric, stochastic semi-supervised with random labeling and \emph{max} combining function (MF-SR) returns the best result with a detection improved by 3.21\%.
The strategy also performs well in terms of AUC-ROC, but is outperformed (by a small margin) by the logarithmic combining function when considering AUC-PR as the performance metric. 

Similarly to the results obtained at the transaction level, the best strategies are those combining the baseline High Risk Querying with stochastic semi-supervised with random labeling (SR) or uncertainty labeling (SU). The addition of an exploratory part (QFU, or SR-U strategies) did not allow to improve the detection accuracy. The performances are even slightly decreased in terms of Top100 Precision for SR-U strategies. 

The results show that the combining function plays an important role. While the \emph{max} and \emph{logarithmic} performed best overall, the \emph{softmax} clearly hampered the fraud detection accuracy. 

\begin{figure*}[!t]
\centering
\subfloat[]{\includegraphics[width=3in, height=3.1in]{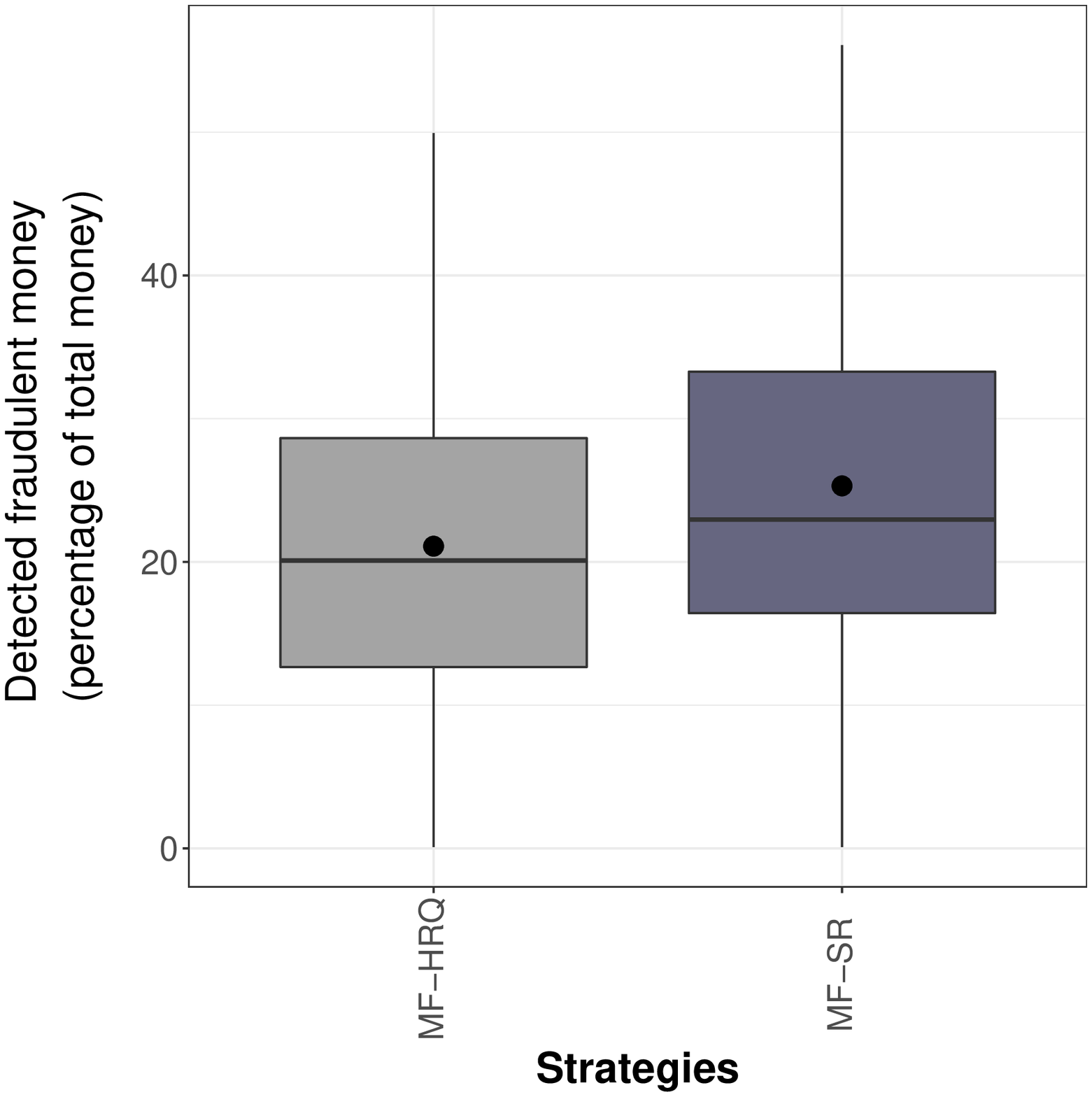}
\label{fig:involved}}
\subfloat[]{\includegraphics[width=3in, height=3.1in]{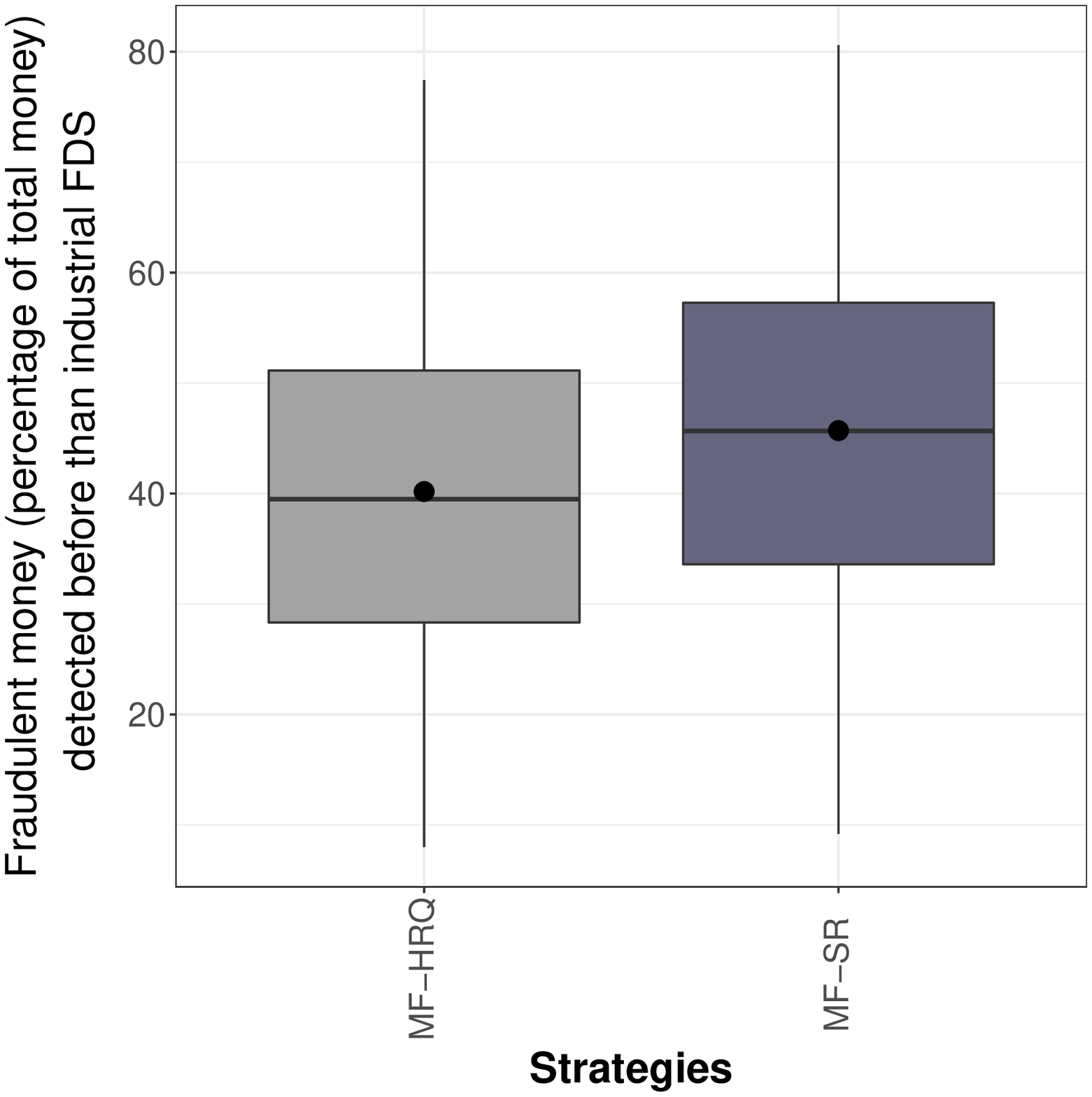}
\label{fig:saved}}
\hfil
\caption{Card-based case study. Daily amount of detected fraudulent money (percentage of daily transacted money) (a). Daily amount of fraudulent money detected before than the industrial FDS (b). Dark boxes indicate the significantly better strategy (paired Wilcoxon test).}\label{fig:find}
\end{figure*}

In Fig.~\ref{fig:find} we show the impact on money savings related to the adoption of the MF-HRQ and MF-SR techniques respectively.
Fig.~\ref{fig:involved} reports the daily amount of fraudulent money detected by two approaches (as a percentage of daily transacted money).
Fig.~\ref{fig:saved} reports the daily amount of fraudulent money which would have been saved if the AL technique would have been used instead of the FDS implemented by our industrial partner (as a percentage of daily fraudulent transacted money).
However, we are not able to measure the amount of fraudulent transactions detected later than our partner since from the historical data we cannot reconstruct if  the fraud was indeed detected by the industrial FDS (or labeled after a claim of the cardholder).

To conclude, stochastic semi-supervised labeling (SR and SU) combined with HRQ remained the best strategies, confirming the results obtained at the transaction level.
Regarding the combining functions, the \emph{logarithmic} function provided the best performance in terms of AUC-ROC, but the \emph{max} function significantly outperformed in terms of Top100 Precision. Overall, the \emph{max} combining function was observed to provide the most stable improvement throughout the range of explored performance metrics. 




\section{Conclusion and future work}

This paper investigates the impact of active learning strategies on the fraud detection accuracy in credit cards.
Using a real-world dataset of several millions of transactions over sixty days, we provided an extensive analysis and comparison of different strategies, involving standard active learning, exploratory active learning, semi-supervised learning and combining functions\footnote{We made the StreamingActiveLearningStrategies repository available   in \url{github.com/fabriziocarcillo/}}.
In particular we assessed the importance of different selection criteria (supervised, unsupervised or semi-supervised). 
Moreover, we provided a two-dimensional visualization of the complexity (e.g. non separability)  of the fraud detection problem as well as a visualization of the distribution of the query points issued by the different strategies.

Our results show that unsupervised outlier detection has very low accuracy in the context of credit card fraud detection.
This is especially true for a context where only a very small number of cards can be investigated every day and high precision is required.

Furthermore we find that the baseline active learning for fraud detection, the Highest Risk Querying, can be noticeably improved by combining it with Stochastic Semi-supervised Learning, thereby allowing to increase the fraud detection accuracy by up to five percent.
Exploratory active learning techniques (including supervised or unsupervised selection criteria) have not improved the fraud detection task.
This can be attributed to the highly imbalanced nature of the data and the small exploration budget that can be reasonably allocated in a fraud detection system.

Last, our results on combining functions for bags of transactions showed that the baseline strategy, implemented with the \emph{max} strategy, was the most stable across different accuracy metrics, but that alternative functions could be worth considering.

Future work will focus on two main directions, the stratification of the dataset according to the properties of transactions or cardholders, the design and assessment of hybrid strategies to combine unsupervised (notably outlier detection) and supervised methods. 

\begin{acknowledgements}
The authors FC, YLB and GB acknowledge the funding of the  Brufence project (Scalable machine learning for automating defense system) supported by INNOVIRIS (Brussels Institute for the encouragement of scientific research and innovation).

Computational resources have been provided by the Consortium des Équipements de Calcul Intensif (CÉCI), funded by the Fonds de la Recherche Scientifique de Belgique (F.R.S.-FNRS) under Grant No. 2.5020.11.

This is a pre-print of an article published in the International Journal of Data Science and Analytics. The final authenticated version is available online at: https://doi.org/[DOI: 10.1007/s41060-018-0116-z]
\\
\\
\\
On behalf of all authors, the corresponding author states that there is no conflict of interest.
\end{acknowledgements}

\end{sloppypar}
\bibliographystyle{spmpsci}      
\bibliography{refs}   

\begin{thebibliography}{10}
\providecommand{\url}[1]{{#1}}
\providecommand{\urlprefix}{URL }
\expandafter\ifx\csname urlstyle\endcsname\relax
  \providecommand{\doi}[1]{DOI~\discretionary{}{}{}#1}\else
  \providecommand{\doi}{DOI~\discretionary{}{}{}\begingroup
  \urlstyle{rm}\Url}\fi

\bibitem{aggarwal2015outlier}
Aggarwal, C.C.: Outlier analysis.
\newblock In: Data mining, pp. 237--263. Springer (2015)

\bibitem{bhattacharyya2011data}
Bhattacharyya, S., Jha, S., Tharakunnel, K., Westland, J.C.: Data mining for
  credit card fraud: A comparative study.
\newblock Decision Support Systems \textbf{50}(3), 602--613 (2011)

\bibitem{bolton2001unsupervised}
Bolton, R.J., Hand, D.J., et~al.: Unsupervised profiling methods for fraud
  detection.
\newblock Credit Scoring and Credit Control VII pp. 235--255 (2001)

\bibitem{breunig2000lof}
Breunig, M.M., Kriegel, H.P., Ng, R.T., Sander, J.: Lof: identifying
  density-based local outliers.
\newblock In: ACM sigmod record, vol.~29, pp. 93--104. ACM (2000)

\bibitem{carcillo2017InfFus}
Carcillo, F., Dal~Pozzolo, A., Le~Borgne, Y.A., Caelen, O., Mazzer, Y.,
  Bontempi, G.: Scarff: a scalable framework for streaming credit card fraud
  detection with spark.
\newblock Information fusion \textbf{41}, 182--194 (2018)

\bibitem{carcillo2017active}
Carcillo, F., Le~Borgne, Y.A., Caelen, O., Bontempi, G.: An assessment of
  streaming active learning strategies for real-life credit card fraud
  detection.
\newblock In: DSAA-The 4th IEEE International Conference on Data Science and
  Advanced Analytics, vol.~7, pp. 783--790 (2017)

\bibitem{chandola2009anomaly}
Chandola, V., Banerjee, A., Kumar, V.: Anomaly detection: A survey.
\newblock ACM computing surveys (CSUR) \textbf{41}(3), 15 (2009)

\bibitem{chapelle2009semi}
Chapelle, O., Scholkopf, B., Zien, A.: Semi-supervised learning (chapelle, o.
  et al., eds.; 2006)[book reviews].
\newblock IEEE Transactions on Neural Networks \textbf{20}(3), 542--542 (2009)

\bibitem{chawlan2002smote}
Chawla, N.V., Bowyer, K.W., Hall, L.O., Kegelmeyer, W.P.: Smote: synthetic
  minority over-sampling technique.
\newblock Journal of artificial intelligence research \textbf{16}, 321--357
  (2002)

\bibitem{chen2004using}
Chen, C., Liaw, A., Breiman, L.: Using random forest to learn imbalanced data.
\newblock University of California, Berkeley \textbf{110} (2004)

\bibitem{cohn1994improving}
Cohn, D., Atlas, L., Ladner, R.: Improving generalization with active learning.
\newblock Machine learning \textbf{15}(2), 201--221 (1994)

\bibitem{pinto2011weighted}
Pinto~da Costa, J.F., Alonso, H., Roque, L.: A weighted principal component
  analysis and its application to gene expression data.
\newblock IEEE/ACM Transactions on Computational Biology and Bioinformatics
  (TCBB) \textbf{8}(1), 246--252 (2011)

\bibitem{dal2014learned}
Dal~Pozzolo, A., Caelen, O., Le~Borgne, Y.A., Waterschoot, S., Bontempi, G.:
  Learned lessons in credit card fraud detection from a practitioner
  perspective.
\newblock Expert systems with applications \textbf{41}(10), 4915--4928 (2014)

\bibitem{dasgupta2011two}
Dasgupta, S.: Two faces of active learning.
\newblock Theoretical computer science \textbf{412}(19), 1767--1781 (2011)

\bibitem{dorronsoro1997neural}
Dorronsoro, J.R., Ginel, F., Sgnchez, C., Cruz, C.: Neural fraud detection in
  credit card operations.
\newblock IEEE transactions on neural networks \textbf{8}(4), 827--834 (1997)

\bibitem{drews2013novelty}
Drews, P., N{\'u}{\~n}ez, P., Rocha, R.P., Campos, M., Dias, J.: Novelty
  detection and segmentation based on gaussian mixture models: A case study in
  3d robotic laser mapping.
\newblock Robotics and Autonomous Systems \textbf{61}(12), 1696--1709 (2013)

\bibitem{ertekin2007learning}
Ertekin, S., Huang, J., Bottou, L., Giles, L.: Learning on the border: active
  learning in imbalanced data classification.
\newblock In: Proceedings of the sixteenth ACM conference on Conference on
  information and knowledge management, pp. 127--136. ACM (2007)

\bibitem{fan2004active}
Fan, W., Huang, Y.a., Wang, H., Yu, P.S.: Active mining of data streams.
\newblock In: Proceedings of the 2004 SIAM International Conference on Data
  Mining, pp. 457--461. SIAM (2004)

\bibitem{ilonen2006gaussian}
Ilonen, J., Paalanen, P., Kamarainen, J.K., Kalviainen, H.: Gaussian mixture
  pdf in one-class classification: computing and utilizing confidence values.
\newblock In: Pattern Recognition, 2006. ICPR 2006. 18th International
  Conference on, vol.~2, pp. 577--580. IEEE (2006)

\bibitem{jacobusse2016selection}
Jacobusse, G., Veenman, C.: On selection bias with imbalanced classes.
\newblock In: International Conference on Discovery Science, pp. 325--340.
  Springer (2016)

\bibitem{japkowicz2002class}
Japkowicz, N., Stephen, S.: The class imbalance problem: A systematic study.
\newblock Intelligent data analysis \textbf{6}(5), 429--449 (2002)

\bibitem{jurgovsky2018sequence}
Jurgovsky, J., Granitzer, M., Ziegler, K., Calabretto, S., Portier, P.E.,
  He-Guelton, L., Caelen, O.: Sequence classification for credit-card fraud
  detection.
\newblock Expert Systems with Applications  (2018)

\bibitem{kriegel2009loop}
Kriegel, H.P., Kr\"{o}ger, P., Schubert, E., Zimek, A.: Loop: Local outlier
  probabilities.
\newblock In: Proceedings of the 18th ACM Conference on Information and
  Knowledge Management, CIKM '09, pp. 1649--1652. ACM, New York, NY, USA
  (2009).
\newblock \doi{10.1145/1645953.1646195}.
\newblock \urlprefix\url{http://doi.acm.org/10.1145/1645953.1646195}

\bibitem{Lewis94heterogeneousuncertainty}
Lewis, D.D., Catlett, J.: Heterogeneous uncertainty sampling for supervised
  learning.
\newblock In: In Proceedings of the Eleventh International Conference on
  Machine Learning, pp. 148--156. Morgan Kaufmann (1994)

\bibitem{lewis1994sequential}
Lewis, D.D., Gale, W.A.: A sequential algorithm for training text classifiers.
\newblock In: Proceedings of the 17th annual international ACM SIGIR conference
  on Research and development in information retrieval, pp. 3--12.
  Springer-Verlag New York, Inc. (1994)

\bibitem{li2016anomaly}
Li, L., Hansman, R.J., Palacios, R., Welsch, R.: Anomaly detection via a
  gaussian mixture model for flight operation and safety monitoring.
\newblock Transportation Research Part C: Emerging Technologies \textbf{64},
  45--57 (2016)

\bibitem{liu2008isolation}
Liu, F.T., Ting, K.M., Zhou, Z.H.: Isolation forest.
\newblock In: Data Mining, 2008. ICDM'08. Eighth IEEE International Conference
  on, pp. 413--422. IEEE (2008)

\bibitem{palau2011burst}
Palau, C., Arregui, F., Carlos, M.: Burst detection in water networks using
  principal component analysis.
\newblock Journal of Water Resources Planning and Management \textbf{138}(1),
  47--54 (2011)

\bibitem{pang2016unsupervised}
Pang, G., Cao, L., Chen, L., Liu, H.: Unsupervised feature selection for
  outlier detection by modelling hierarchical value-feature couplings.
\newblock In: Data Mining (ICDM), 2016 IEEE 16th International Conference on,
  pp. 410--419. IEEE (2016)

\bibitem{pang2017learning}
Pang, G., Cao, L., Chen, L., Liu, H.: Learning homophily couplings from non-iid
  data for joint feature selection and noise-resilient outlier detection.
\newblock In: Proceedings of the 26th International Joint Conference on
  Artificial Intelligence, pp. 2585--2591. AAAI Press (2017)

\bibitem{pichara2008detection}
Pichara, K., Soto, A., Araneda, A.: Detection of anomalies in large datasets
  using an active learning scheme based on dirichlet distributions.
\newblock In: Ibero-American Conference on Artificial Intelligence, pp.
  163--172. Springer (2008)

\bibitem{pimentel2014review}
Pimentel, M.A., Clifton, D.A., Clifton, L., Tarassenko, L.: A review of novelty
  detection.
\newblock Signal Processing \textbf{99}, 215--249 (2014)

\bibitem{dal2017TNNLS}
Pozzolo, A.D., Boracchi, G., Caelen, O., Alippi, C., Bontempi, G.: Credit card
  fraud detection: A realistic modeling and a novel learning strategy.
\newblock IEEE Transactions on Neural Networks and Learning Systems
  \textbf{PP}(99), 1--14 (2018).
\newblock \doi{10.1109/TNNLS.2017.2736643}

\bibitem{ren2004rdf}
Ren, D., Wang, B., Perrizo, W.: Rdf: A density-based outlier detection method
  using vertical data representation.
\newblock In: Data Mining, 2004. ICDM'04. Fourth IEEE International Conference
  on, pp. 503--506. IEEE (2004)

\bibitem{Rokach2016111}
Rokach, L.: Decision forest: Twenty years of research.
\newblock Information Fusion \textbf{27}, 111--125 (2016)

\bibitem{sahin2013cost}
Sahin, Y., Bulkan, S., Duman, E.: A cost-sensitive decision tree approach for
  fraud detection.
\newblock Expert Systems with Applications \textbf{40}(15), 5916--5923 (2013)

\bibitem{schohn2000less}
Schohn, G., Cohn, D.: Less is more: Active learning with support vector
  machines.
\newblock In: ICML, pp. 839--846. Citeseer (2000)

\bibitem{scholkopf2000support}
Sch{\"o}lkopf, B., Williamson, R.C., Smola, A.J., Shawe-Taylor, J., Platt,
  J.C.: Support vector method for novelty detection.
\newblock In: Advances in neural information processing systems, pp. 582--588
  (2000)

\bibitem{seeja2014fraudminer}
Seeja, K., Zareapoor, M.: Fraudminer: a novel credit card fraud detection model
  based on frequent itemset mining.
\newblock The Scientific World Journal \textbf{2014}(0), 1--10 (2014)

\bibitem{sethi2014revived}
Sethi, N., Gera, A.: A revived survey of various credit card fraud detection
  techniques.
\newblock International Journal of Computer Science and Mobile Computing
  \textbf{3}(4), 780--791 (2014)

\bibitem{settles2010active}
Settles, B.: Active learning literature survey.
\newblock University of Wisconsin, Madison \textbf{52}(55-66), 11 (2010)

\bibitem{settles2008multiple}
Settles, B., Craven, M., Ray, S.: Multiple-instance active learning.
\newblock In: Advances in neural information processing systems, pp. 1289--1296
  (2008)

\bibitem{shimpi2015survey}
Shimpi, P.R., Kadroli, V.: Survey on credit card fraud detection techniques.
\newblock International Journal Of Engineering And Computer Science
  \textbf{4}(11), 15,010--15,015 (2015)

\bibitem{shyu2003novel}
Shyu, M.L., Chen, S.C., Sarinnapakorn, K., Chang, L.: A novel anomaly detection
  scheme based on principal component classifier.
\newblock Tech. rep., MIAMI UNIV CORAL GABLES FL DEPT OF ELECTRICAL AND
  COMPUTER ENGINEERING (2003)

\bibitem{srivastava2008credit}
Srivastava, A., Kundu, A., Sural, S., Majumdar, A.: Credit card fraud detection
  using hidden markov model.
\newblock IEEE Transactions on dependable and secure computing \textbf{5}(1),
  37--48 (2008)

\bibitem{tang2002enhancing}
Tang, J., Chen, Z., Fu, A., Cheung, D.: Enhancing effectiveness of outlier
  detections for low density patterns.
\newblock Advances in Knowledge Discovery and Data Mining \textbf{0}, 535--548
  (2002)

\bibitem{van2015apate}
Van~Vlasselaer, V., Bravo, C., Caelen, O., Eliassi-Rad, T., Akoglu, L., Snoeck,
  M., Baesens, B.: Apate: A novel approach for automated credit card
  transaction fraud detection using network-based extensions.
\newblock Decision Support Systems \textbf{75}, 38--48 (2015)

\bibitem{van2015afraid}
Van~Vlasselaer, V., Eliassi-Rad, T., Akoglu, L., Snoeck, M., Baesens, B.:
  Afraid: fraud detection via active inference in time-evolving social
  networks.
\newblock In: Advances in Social Networks Analysis and Mining (ASONAM), 2015
  IEEE/ACM International Conference on, pp. 659--666. IEEE (2015)

\bibitem{vijayanarasimhan2010far}
Vijayanarasimhan, S., Jain, P., Grauman, K.: Far-sighted active learning on a
  budget for image and video recognition.
\newblock In: Computer Vision and Pattern Recognition (CVPR), 2010 IEEE
  Conference on, pp. 3035--3042. IEEE (2010)

\bibitem{wang2004novel}
Wang, W., Guan, X., Zhang, X.: A novel intrusion detection method based on
  principle component analysis in computer security.
\newblock Advances in Neural Networks-ISNN 2004 pp. 88--89 (2004)

\bibitem{wei2013effective}
Wei, W., Li, J., Cao, L., Ou, Y., Chen, J.: Effective detection of
  sophisticated online banking fraud on extremely imbalanced data.
\newblock World Wide Web \textbf{16}(4), 449--475 (2013)

\bibitem{xie2011stochastic}
Xie, J., Xiong, T.: Stochastic semi-supervised learning on partially labeled
  imbalanced data.
\newblock In: Active Learning and Experimental Design workshop In conjunction
  with AISTATS 2010, pp. 85--98. AISTATS (2011)

\bibitem{zareapoor2015application}
Zareapoor, M., Shamsolmoali, P.: Application of credit card fraud detection:
  Based on bagging ensemble classifier.
\newblock Procedia Computer Science \textbf{48}, 679--685 (2015)

\bibitem{zhang2009new}
Zhang, K., Hutter, M., Jin, H.: A new local distance-based outlier detection
  approach for scattered real-world data.
\newblock Advances in knowledge discovery and data mining \textbf{5476}

\bibitem{zhang2017detection}
Zhang, Y., Bingham, C., Mart{\'\i}nez-Garc{\'\i}a, M., Cox, D.: Detection of
  emerging faults on industrial gas turbines using extended gaussian mixture
  models.
\newblock International Journal of Rotating Machinery \textbf{2017}, 1--9
  (2017)

\bibitem{zhu2007active}
Zhu, J., Hovy, E.H.: Active learning for word sense disambiguation with methods
  for addressing the class imbalance problem.
\newblock In: EMNLP-CoNLL, vol.~7, pp. 783--790 (2007)

\bibitem{vzliobaite2011active}
{\v{Z}}liobaite, I., Bifet, A., Pfahringer, B., Holmes, G.: Active learning
  with evolving streaming data.
\newblock In: Joint European Conference on Machine Learning and Knowledge
  Discovery in Databases, pp. 597--612. Springer (2011)

\end{thebibliography}

%
%

\end{document}